\documentclass[lettersize,journal]{IEEEtran}
\usepackage{amsmath}
\usepackage{mathtools}
\usepackage[english]{babel}
\usepackage[nottoc]{tocbibind}

\usepackage{amssymb}
\usepackage{algorithm}
\usepackage{algpseudocode}
\usepackage{accents}
\usepackage{multirow}
\usepackage{cite} 

\usepackage{subcaption}
\usepackage{caption}

\usepackage{tikz}

\usepackage{color}
\makeatletter
\def\zz{\edef\zzz{\pdfliteral{\current@color}}}
\makeatother

\usepackage{graphicx}

\newcounter{theorem}

\newtheorem{definition}[theorem]{Definition}
\DeclareMathOperator*{\argmin}{arg\,min}
\allowdisplaybreaks
\usepackage{moreverb,url}

\newcommand{\der}{\mathrm{D}}
\usepackage[colorlinks,bookmarksopen,bookmarksnumbered,citecolor=red,urlcolor=red]{hyperref}

\def\volumeyear{2022}

\newcommand\copyrighttext{%
 \textcopyright 2023 IEEE. Personal use of this material is permitted.
  Permission from IEEE must be obtained for all other uses, in any current or future
  media, including reprinting/republishing this material for advertising or promotional
  purposes, creating new collective works, for resale or redistribution to servers or
  lists, or reuse of any copyrighted component of this work in other works.
  DOI: \href{https://doi.org/10.1109/TRO.2023.3308773}{10.1109/TRO.2023.3308773}}
\newcommand\copyrightnotice{%
\begin{tikzpicture}[remember picture,overlay]
\node[anchor=south,yshift=5pt] at (current page.south) {\fbox{\parbox{\dimexpr\textwidth-\fboxsep-\fboxrule\relax}{  \footnotesize \copyrighttext}}};
\end{tikzpicture}%
}

\begin{document}

\title{Hybrid iLQR Model Predictive Control for Contact Implicit Stabilization on Legged Robots}

\author{Nathan J. Kong$^{1{,2}}$, Chuanzheng Li$^{2}$, {George Council$^{1}$}, and Aaron M. Johnson$^{1}$%
\thanks{This material is based upon work supported by the U.S. Army Research Office under grant \#W911NF-19-1-0080, the National Science Foundation under grant \#ECCS-1924723, and XPENG Robotics. The views and conclusions contained in this document are those of the authors and should not be interpreted as representing the official policies, either expressed or implied, of the Army Research Office, National Science Foundation, or the U.S. Government. The U.S. Government is authorized to reproduce and distribute reprints for Government purposes notwithstanding any copyright notation herein. 
Corresponding author {A.~M.~Johnson} {\tt\small {amj1@andrew.cmu.edu}}.}%
\thanks{$^{1}$ Department of Mechanical Engineering, Carnegie Mellon University, Pittsburgh, Pennsylvania, 15213}
\thanks{$^{2}$ XPENG Robotics, Mountain View, California, 95054}
}

\thispagestyle{empty}
\setcounter{page}{0}
\begin{figure*}[t!]
\centering
\large
This paper has been accepted for publication in IEEE Transactions on Robotics.\\

DOI: \href{https://doi.org/10.1109/TRO.2023.3308773}{10.1109/TRO.2023.3308773}\\

IEEE Explore: \href{https://ieeexplore.ieee.org/document/10252162}{https://ieeexplore.ieee.org/document/10252162}\\

~\\

Please cite the paper as:\\

Nathan J. Kong, Chuanzheng Li, George Council, and Aaron M. Johnson, ``Hybrid iLQR Model Predictive Control for Contact Implicit Stabilization on Legged Robots,'' in \emph{IEEE Transactions on Robotics}, 2023.\\

~\\

~\\

\copyrighttext
\vspace{400px}
\end{figure*}
\maketitle
\copyrightnotice





\begin{abstract}
Model Predictive Control (MPC) is a popular strategy for controlling robots but is difficult for systems with contact due to the complex nature of hybrid dynamics.
To implement MPC for systems with contact, dynamic models are often simplified or contact sequences fixed in time in order to plan trajectories efficiently.
In this work, we {propose the} Hybrid iterative Linear Quadratic Regulator
{(HiLQR), which extends iLQR to a class of piecewisesmooth
hybrid dynamical systems with state jumps. This is accomplished by 1) allowing
for changing hybrid modes in the forward pass, 2) using the
saltation matrix to update the gradient information in the
backwards pass, and 3) using a reference extension to account for
mode mismatch.
We demonstrate these changes on a variety
of hybrid systems and compare the different strategies for
computing the gradients.
We further show how HiLQR can} work in a MPC fashion (HiLQR MPC) by 1) modifying how the cost function is computed when contact modes do not align, 2) utilizing parallelizations when simulating rigid body dynamics, and 3) using efficient analytical derivative computations of the rigid body dynamics.
The result is a system that can modify the contact sequence of the reference behavior and plan whole body motions cohesively -- which is crucial when dealing with large perturbations.
HiLQR MPC is tested on two systems: first, the hybrid cost modification is validated on a simple actuated bouncing ball hybrid system.
Then HiLQR MPC is compared against methods that utilize centroidal dynamic assumptions on a quadruped robot (Unitree A1).
HiLQR MPC outperforms the centroidal methods in both simulation and hardware tests.
\end{abstract}


\section{Introduction}
\begin{figure*}[htbp]
    \centering
    \includegraphics[width=\textwidth]{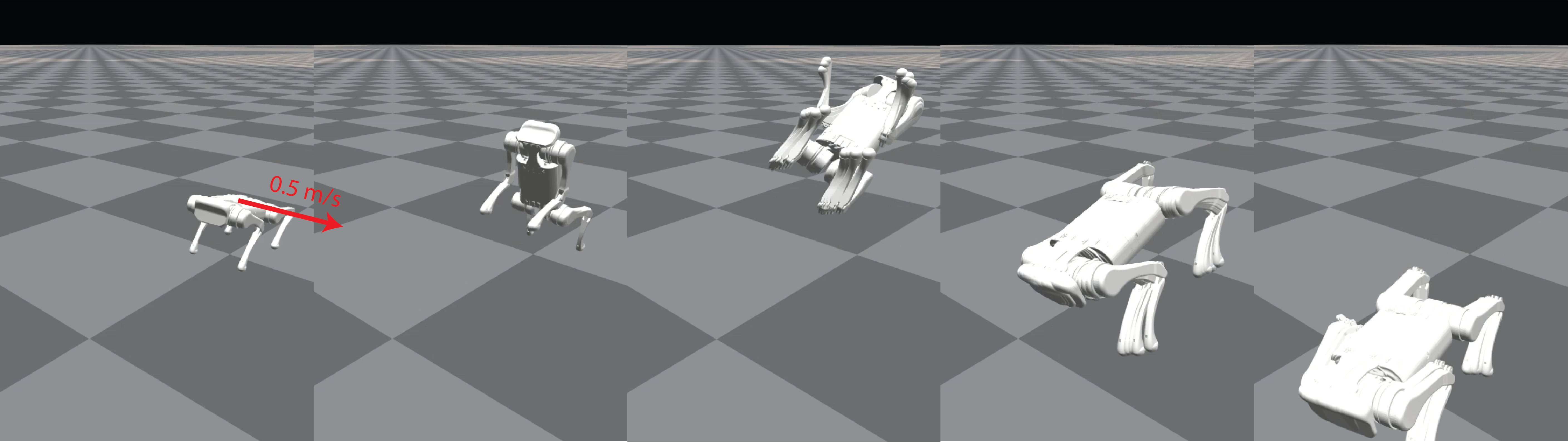}
    \caption{HiLQR MPC forward pass for tracking a backflip with an initial 0.5 m/s lateral perturbation on the body. 
    9 robot models are used on the forward pass to solve the line search in parallel.}
    \label{fig:backflip}
\end{figure*}
In order for robots to reliably move and interact within our unstructured world, they need to be able to replan motions to handle unexpected perturbations or changes in the environment.
However, replanning is difficult for robotic systems that have changing contact with the world because of the complexity of the discontinuous dynamics and combinatoric issues that arise.

There are many methods for planning contact-rich behaviors offline \cite{posa2014direct,mordatch2012discovery,mombaur2009using,diehl2006fast}, but these methods generally suffer from poor time complexity and cannot be used directly in real-time applications.
Direct methods for contact implicit trajectory optimization \cite{posa2014direct,mordatch2012discovery} simultaneously solve for the states, inputs, and contact forces of an optimal trajectory while encoding the contact conditions through complementarity constraints -- which are notoriously difficult and slow to solve.
A relaxation of contact implicit trajectory optimization is to fix the contact sequence for each timestep \cite{von1999user,posa2016optimization,kelly2017introduction,pardo2017hybrid,winkler2018gait}.

{
To allow efficient updates of the contact sequence, \cite{manchester2021fast} speeds up contact implicit trajectory optimization through strategic linearization about a target trajectory.
However, the basin of attraction is smaller because it is linearized about a single nominal trajectory. 
If the robot needs to drastically change the trajectory, the controller will not use a good model given the linearization of the target trajectory.}

Other relaxations have been made for the planning problem to achieve real-time Model Predictive Control (MPC). 
Centroidal methods \cite{di2018dynamic,kim2019highly,da2020learning,xie2021glide,gehring2013control} have had a lot of success in planning gaits in real-time by making large simplifications on the robot dynamics and also assuming a fixed contact sequence.
Swing legs are often controlled separately using Raibert heuristics \cite{raibert1986legged} {or} capture point methods \cite{pratt2006capture} to regulate body {velocity}.
However, simplifications to the robot dynamics can lead to the controller being less robust to perturbations which require reasoning about the full dynamics, such as nonlinear changes in lever arm for leg extension, varying inertia when the leg changes shape, or not accounting for  changes in contact.

Shooting methods which utilize Differential Dynamic Programming (DDP) \cite{mayne1966second} or iterative Linear Quadratic Regulator (iLQR) {\cite{li2004iterative,tassa2012synthesis,koenemann2015whole,neunert2018whole,dantec2021whole}} are good candidates for model predictive control because they are fast, can utilize the full nonlinear dynamics, and solutions are always dynamically feasible.
Methods that utilize the full nonlinear dynamics {\cite{li2020hybrid,mastalli2020crocoddyl,mastalli2022agile}} generally come at the cost of enforcing a fixed contact sequence.
\cite{li2021model} utilizes the full nonlinear dynamics for timesteps closer to the current horizon and then uses simplified dynamics for timesteps later in the future, but also uses a fixed contact sequence.
{These methods rely on iterative linearization, but have not previously considered the best way to linearize systems with changing contact conditions.}

{In this work we propose the Hybrid iLQR (HiLQR) trajectory optimization algorithm that extends iLQR to be
a full-order contact implicit trajectory optimization algorithm (Sec.~\ref{sec:HiLQR}).
This is accomplished by:
\begin{enumerate}
    \item Allowing for varying mode sequences on the forward
pass by using event detection to dictate when a transition
occurs and enforcing the appropriate dynamics in
each mode, Sec.~\ref{section:hybrid_forwards}.
    \item Applying the reset map on the forward pass and propagating
the value function through reset maps in the
backwards pass by using a saltation matrix, the linearization of the sensitivity equation for this class of hybrid systems, Sec.~\ref{section:hybrid_backwards}.
    \item Using reference extensions when there is a mode
mismatch to get a valid control input in each mode,
Sec.~\ref{section:hybrid_extensions}.
\end{enumerate}}
Previous attempts to apply DDP/iLQR to hybrid systems, e.g. \cite{li2020hybrid}, utilized a prespecified hybrid mode sequence and used the Jacobian of the reset map to approximate the value function through a hybrid transition.
In contrast, here we use the saltation matrix (Def.~\ref{def:salt}), \cite{leine2004dynamics, rijnen2015optimal, aizerman1958determination, burden2018contraction}, to propagate the value function in the backwards pass. 
This change makes a significant difference in solution quality and convergence, as we show in Sec.~\ref{sec:experiment-trajopt}. Furthermore, the switching constraints are enforced as part of the dynamics on the forward pass -- if the current timestep reaches a hybrid event, the solution jumps to the next hybrid mode using the reset map. These changes enable HiLQR to be run as a standalone algorithm with improved solution quality and convergence properties.

This manuscript is evolved from \cite{paper:kong-ilqr-2021}, which presented the initial HiLQR algorithm. In this extension, we show how HiLQR can be used as an online Model Predictive Control (MPC) controller (Sec.~\ref{sec:HiLQR_MPC}) and present new results demonstrating the efficacy both in simulation and on a physical robot (Sec.~\ref{sec:mpc_experiments}). 
This is accomplished by adapting the cost function to include a hybrid cost update as shown in Sec. \ref{subsec:hybridcost}, creating an optimization framework which utilizes fast parallel simulations of full order legged robots as shown in Sec. \ref{subsec:rollout}, and applying the saltation matrix to fast analytical derivative calculations as shown in Sec. \ref{subsec:backward}.
Through these changes,\ Hybrid iLQR MPC {can greatly modify} the contact sequence when stabilizing large perturbations, e.g.\ as shown in Fig.\ \ref{fig:backflip}, {because the optimizer is not constrained to the reference trajectory's gait sequence}.
We show that HiLQR MPC can reject bigger disturbances than centroidal methods when perturbed along a walking trajectory {(Sec.~\ref{sec:simulated})}.
We also show that HiLQR MPC working on a real robot in real-time can reject disturbances more reliably than centroidal methods {(Sec.~\ref{sec:physical})}.

\section{Hybrid systems background}
In this section, we define what a hybrid system is and the first order linearization for hybrid events. 
Also, two different hybrid simulation techniques are reviewed for rigid body systems with unilateral constraints.
\subsection{Hybrid Systems}
This section closely follows the formulation of hybrid systems from \cite{johnson2016hybrid}.
\begin{definition} \label{def:hs}
    A \textbf{hybrid dynamical system} \cite{Back_Guckenheimer_Myers_1993,lygeros2003dynamical,goebel2009hybrid} is a tuple $\mathcal{H} := (\mathcal{J},{\mathnormal{\Gamma}},\mathcal{D},\mathcal{F},\mathcal{G},\mathcal{R})$ where the parts are defined as:
    \begin{enumerate}
        \item $\mathcal{J} := \{I,J,...,K\} \subset \mathbb{N}$ is the finite set of discrete \textbf{modes}.
        \item $\mathnormal{\Gamma} \subset \mathcal{J}\times\mathcal{J}$ is the set of discrete \textbf{transitions} that form a directed graph structure on $\mathcal{J}$.
        \item $\mathcal{D}$ is the collection of \textbf{domains} $D_I$.
        \item $\mathcal{F}$ is a collection of time-varying \textbf{vector fields} $F_I$.
        \item $\mathcal{G}$ is the collection of \textbf{guards} where $G_{(I,J)}(t)= \{(x,t) \in D_I|g_{(I,J)}(t,x)\leq0\}$.
        \item $\mathcal{R}$ is called the \textbf{reset} that maps the state from $D_I$ to $D_J$ when the guard $G_{(I,J)}$ is met.
    \end{enumerate}
    \end{definition}
An example hybrid execution may consist of a starting point $x_0$ in ${D}_I$ flowing with dynamics ${F}_I$ and reaching the guard condition $g_{(I,J)}(x,t) = 0$, applying the reset map $R_{(I,J)}(x,t)$ resetting into ${D}_J$ and then flowing with the new dynamics ${F}_J$. 

\begin{definition} \label{def:salt}
    The \textbf{saltation matrix}  \cite{aizerman1958determination,leine2004dynamics, rijnen2015optimal,burden2018contraction}
    \begin{equation}
        \Xi := {\der}_x R+\frac{\left(F_J-{\der}_xR \cdot F_I - {\der}_tR\right) {\der}_x g}{{\der}_t g +{\der}_x g \cdot F_I} \label{eq:saltationmatrix}
    \end{equation}
    is the first order approximation of the variational update at hybrid transitions from mode $I$ to $J$ evaluated at time $t$, pre-impact state
    $x(t^-)$, and control input $u(t^-)$, with $F_I$ evaluated at $F_I(t^+,x(t^-),u(t^-))$ and $F_J$ evaluated at $F_J(t^+,x(t^+),u(t^-))$, where $x(t^+) = {R}_{(I,J)}(t^-,x(t^-))$, and $\der_x$ is the Jacobian with respect to state and $\der_t$ is the derivative with respect to time.
    It maps perturbations to first order from pre-transition $\delta x(t^-)$ to post-transition $\delta x(t^+)$ in the following way
    \begin{equation}
        \delta x(t^+) = \Xi \delta x(t^-) + \text{h.o.t.}
        \label{eq::saltperturbation}
    \end{equation}
    where \emph{h.o.t.} represents higher order terms.
    \end{definition}
\subsection{Hybrid Simulators}
There are 2 main hybrid simulation techniques for rigid bodies with unilateral constraints -- event-driven and timestepping.
HiLQR MPC uses a hybrid simulator and can use either method.
But different modifications need to be made depending on which simulation type is used.
It is important to have a high level understanding of each of these simulation types to understand that modifications discussed in this work.

Event-driven hybrid simulators \cite{wehage1982dynamic,pfeiffer1996multibody,brogliato2002numerical} follow very closely to the example shown in the definition of hybrid dynamical systems Def. \ref{def:hs}.
Event-driven simulations are convenient because the dynamics have a well defined structure and contacts are persistently maintained.
However, event-driven simulations have problems with behaviors like Zeno {\cite{lygeros2003dynamical}}, where an infinite number of hybrid transitions are made in a finite amount of time, as they must stop integration and apply a reset map for each individual event.

Time-stepping \cite{stewart1996implicit,anitescu1997formulating,brogliato2002numerical} schemes circumvent issues like Zeno by integrating impulses over small timesteps at a time and are numerically efficient, especially for systems with large numbers of constraints.
These methods allow contact constraints to be added or removed at any time step, but only once per time step. 
Furthermore, no distinction is made between continuous contact forces and discontinuous impulses. However, they are limited to first-order (Euler) integration of the dynamics.

{Time-stepping methods are commonly employed for simulations involving hard contacts since they avoid the Zeno problem and simplify the process of verifying all potential contact modes. On the other hand, event-driven hybrid simulators are capable of modeling a broader range of hybrid systems and are not limited to computing the progression of constrained systems.}
\section{Hybrid iLQR}
\label{sec:HiLQR}

This section 
covers  an abridged derivation of iLQR \cite{li2004iterative} following \cite{tassa2012synthesis}, proposes the changes to make iLQR work on hybrid systems, and discusses several important key features of the new algorithm.
 \subsection{Smooth iLQR background}
Consider a nonlinear dynamical system with states $x\in\mathbb{R}^n$, inputs $u\in \mathbb{R}^m$, and dynamics $\dot{x} = F(x(t),u(t))$.
Define a discretization of the continuous dynamics over a timestep $\Delta$ such that at time $t_k$ the discrete dynamics are $x_{k+1} = f_\Delta(x_k,u_k)$, 
where $t_{k+1} =t_k+\Delta$, $x_k = x(t_k)$, and $u_k = u(t_k)$. 
Let $U := \{u_0,u_1,...,u_{N-1}\}$ be the input sequence, $J_N$ the terminal cost, and $J$ the runtime cost, where $J$ and $J_N$ are both differentiable functions into $\mathbb{R}$.

The optimal control problem over $N$ timesteps is
\begin{align} 
\min_{U} \quad & J_N(x_N) + 
      \sum_{i=0}^{N-1} J(x_i,u_i) \\ 
\text{where} \quad & x_0 = x(0)\\
& x_{i+1} = f_\Delta(x_i,u_i) \quad \forall i \in \{0, ..., N-1\} \label{eq:dynamics_constraint}
\end{align}

To solve this problem, DDP/iLQR uses Bellman recursion to find the optimal input sequence $U$, we which briefly review here.
Let $U_k:=\{u_k,u_{k+1},...,u_{N-1} \}$ be the sequence of inputs including and after timestep $k$. Define the cost-to-go $J_k$ as the cost incurred including and after timestep $k$
\begin{align}
    J_k(x_k,U_k): = J_N(x_N) + \sum_{i=k}^{N-1} J(x_i,u_i)  \label{eq:costfn}
\end{align}
with $\{x_{k+1},...,x_N\}$ the sequence of states starting at $x_k$ based on $U_k$ and \eqref{eq:dynamics_constraint}.
The value function $V(x,k)$ (Bellman equation), evaluated at state and time $(x_k,k)$ is the optimal cost to go $J_k(x_k,U_k)$,
which can be rewritten as a recursive function of variables from the current timestep using the dynamics \eqref{eq:dynamics_constraint},
\begin{align} 
V(x_k, k) := & \min_{u_k} \quad J(x_k, u_k)  + V(f_\Delta(x_k,u_k),k+1) \label{eq:substituted_value}
\end{align}
Since there is no input at the last timestep, the boundary condition of the value is  the terminal cost, $V_N(x_N,N) :=  J_N(x_N)$.
Next, define  $Q_k$ to be the argument optimized in \eqref{eq:substituted_value}.
Optimizing the Bellman equation directly is incredibly difficult. DDP/iLQR uses a second order local approximation of $Q$ where perturbations about the state and input $(x_k, u_k)$ are taken at time $k$. 
The resulting function is defined to be
\begin{align}
    Q_k(\delta x,\delta u, k) := &J(x_k+\delta x, u_k+\delta u)- J(x_k, u_k) \\
    & +V(f_\Delta(x_k+\delta x,u_k+\delta u),k+1) \nonumber\\ 
    & -V(f_\Delta(x_k,u_k),k+1) \nonumber
\end{align}
where the value function expansion is for timestep $k+1$ and when expanded to second order
\begin{align}
    Q(\delta x,\delta u, k) \approx \frac{1}{2}\begin{bmatrix}1\\\delta x\\\delta u\end{bmatrix}^T\begin{bmatrix}
    0&Q_x^T&Q_u^T\\
    Q_x&Q_{xx}&Q_{ux}^T\\
    Q_u&Q_{ux}&Q_{uu}\\
    \end{bmatrix}\begin{bmatrix}1\\\delta x\\\delta u\end{bmatrix}\label{eq:delta_Q}
\end{align}
the expansion coefficients are
\begin{align}
    Q_{x}  &=  J_{x} + f_{x}^T V_{x}\label{eq:expansion_x}\\
    Q_{u}  &=  J_{u} + f_{u}^TV_{x}\label{eq:expansion_u}\\
    Q_{xx} &= J_{xx} + f_{x}^T V_{xx} f_{x} + V_{x} f_{xx}\label{eq:expansion_xx}\\
    Q_{ux} &= J_{ux} + f_{u}^T V_{xx} f_{x}+ V_{x} f_{uu}\label{eq:expansion_ux}\\
    Q_{uu} &= J_{uu} + f_{u}^T V_{xx} f_{u}+ V_{x} f_{ux}\label{eq:expansion_uu}
\end{align}
where subscripted variables represent derivatives of the function with respect to the variable (e.g.\ $J_x = \der_x J$) and the discretized dynamics are abreviated as $f_k=f_\Delta(x_k,u_k)$.
Note that the second derivative terms (where adjacency indicates tensor contraction) with respect to the dynamics ($f_{xx,k}$, $f_{uu,k}$, and $f_{ux,k}$) in \eqref{eq:expansion_xx}--\eqref{eq:expansion_uu} are used in DDP but ignored in iLQR. 

With this value function expansion, the optimal control input, $\delta u^*$, can be found by setting the derivative of $Q(\delta x,\delta u)$ with respect to $\delta u$ to zero and solving for $\delta u$,
\begin{align}
    \delta u^* = & \argmin_{\delta u} Q(\delta x,\delta u) = -Q_{uu}^{-1}(Q_u+Q_{ux}\delta x)
\end{align}
This optimal control input can be split into a feedforward term $u_{ff} = -Q_{uu}^{-1}Q_u$ and a feedback term $K = -Q_{uu}^{-1}Q_{ux}\delta x$. Therefore, the optimal input for the local approximation at timestep $k$ is the sum of the original input and the optimal control input,
$u_k^* = u_k + \delta u^*$.

Once the optimal controller is defined, the expansion coefficients of $V$ for timestep $k$ can be updated by plugging in the optimal controller into \eqref{eq:delta_Q}
\begin{align}
    V_{x} &= Q_{x}-Q_{u}Q_{uu}^{-1}Q_{ux}\\
    V_{xx} &= Q_{xx}-Q_{ux}^TQ_{uu}^{-1}Q_{ux}
\end{align}
Now that the expansion terms for the value function at timestep $k$ can be expressed as sole a function of $k+1$ the optimal control input can be calculated recursively and stored $(u_{ff,k},K_k)$. This process is called the backwards pass.

Once the backwards pass is completed, a forward pass is run by simulating the dynamics given the new gain schedule $(u_{ff,k},K_k)$ and the previous iterations sequence of states and inputs.
\begin{align}
    \hat{x}_0 &= x_0\\
    \hat{u}_k &= K_k(\hat{x}_k - x_k) + \alpha u_{ff,k}\\
   \hat{x}_{k+1} &= f_\Delta(\hat{x}_k,\hat{u}_k) 
\end{align}
where the new trajectory is denoted with hats $(\hat{x},\hat{u})$ and $\alpha$ is used as a backtracking line-search parameters $0<\alpha\leq1$ \cite[Eqn. 12]{tassa2012synthesis}. The backwards and forwards passes are run until convergence. Following \cite{tassa2012synthesis},  convergence is when the magnitude of the total expected reduction $\delta J$ is small
\begin{align}
    \delta J(\alpha) = \alpha\sum^{N-1}_{i=0}u_{ff,i}^TQ_{u,i} + \frac{\alpha}{2}\sum^{N-1}_{k=0}u_{ff,i}^TQ_{uu,i}u_{ff,i}
    \label{eq:expected_reduction}
\end{align}

Convergence issues may occur when $Q_{uu}$ is not positive-definite or when the second order approximations are inaccurate. 
Regularization is often added to address these issues and here we use the standard regularization from~\cite{mayne1973differential} where a scaled diagonal term is added to the local control cost Hessian.

\subsection{Hybrid system modifications to the forward pass}\label{section:hybrid_forwards}

The first change that is required for iLQR to work on hybrid dynamical systems is that the forward pass must accurately generate the hybrid system execution. The dynamics are integrated for the currently active mode $I_j$
for the duration of the hybrid time period $j$, i.e.\ $\forall t \in  [\underbar{$t$}_{j},\bar{t}_{j}]$, until a guard condition is met,

\begin{align}
        g_{(I_{j},I_{j+1})}(\bar{t}_{j},x(\bar{t}_{j}),u(\bar{t}_{j})) = 0
        \label{eq:guardcond}
\end{align}

To capture these hybrid dynamics in the discrete forward pass, the discretized dynamics are computed using numerical integration with event detection, so that if no event occurs the dynamic update, \eqref{eq:dynamics_constraint}, is, 
\begin{align}
    f_{\Delta_j}(\hat{x}_k,\hat{u}_k) := \int_{t_k}^{t_{k+1}} f_{I_j}(x(t),\hat{u}_k)dt + \hat{x}_k
\end{align}
If during the integration the hybrid guard condition is met, \eqref{eq:guardcond}, the integration halts, the transition state is stored, the reset map is applied, and then the integration is continued with the dynamics of the new mode, $I_{j+1}$. Suppose that the guard condition is met once (which is ensured for small times by transversality) at time $\bar{t}_{j}$, such that $t_k \leq \bar{t}_j \leq t_{k+1}$, then 
\begin{align}
    f_{\Delta}'(\hat{x}_k,\hat{u}_k) =&
    \int_{\underbar{$t$}_{j+1}}^{t_{k+1}} f_{I_{j+1}}(x(t),\hat{u}_k)dt+ \label{eq:transition_timestep}\\ 
    & \quad R_{(I_{j},I_{j+1})}\left(\bar{t}_{j},\int_{t_k}^{\bar{t}_{j}} f_{I_j}(x(t),\hat{u}_k)dt +\hat{x}_k\right)\nonumber
\end{align}
Note that this process can be repeated for as finitely many times as there are hybrid changes during a single timestep, but there cannot be infinitely many changes during a single timestep (no Zeno). 

Finally, in addition to updating the dynamics the cost function, \eqref{eq:costfn}, can be augmented with additional cost terms, $J_{N_j}$, associated with each hybrid transition between the $M$ hybrid modes, as shown in \cite{lantoine2012hybrid},
\begin{align}
    J_0 = J_N (x_N) + \sum_{i=0}^{N-1} J(x_i, u_i) + \sum_{j=1}^{M-1} J_{N_j}(x_{N_j}) 
    \label{eq:hybridcost}
\end{align}
Such an addition may be desirable if e.g., one wanted to penalize the occurrences of a transition event in the hopes of having a minimal number of hybrid events.

\subsection{Hybrid system modifications to the backwards pass} \label{section:hybrid_backwards}
The backwards pass must be updated to reflect the discrete jumps that were added through the hybrid transitions. Away from hybrid transitions, the dynamics are smooth and behave the same way as in the smooth iLQR backwards pass, so our modification to the backwards pass occurs at timesteps where a hybrid transition is made. By substituting \eqref{eq:transition_timestep} into \eqref{eq:substituted_value}, and adding the transition cost from \eqref{eq:hybridcost}, the resulting Bellman equation for the timesteps during hybrid transition $j$ over timestep $k$ is
\begin{align} 
V(x_k,k) = & \min_{U_k} J(x_k, u_k)\!+\! J_{N_j}(x_{N_j})\!+\!V(f_{\Delta}'(x_k,u_k),k\!+\!1) \label{eq:transition_bellman}
\end{align}

We elect to approximate the hybrid transition timestep to have the hybrid event occur at the end of the timestep in order to maintain smooth control inputs for each hybrid epoch.
For the backwards pass to work on the Bellman equation during transition timesteps, we need to find the linearization of $f_{\Delta}'(x_k,u_k)$. 
This linearization step is straight forward when using the saltation matrix to map perturbations pre and post hybrid transition \eqref{eq::saltperturbation}. 

The linearization can be broken up into 2 different steps, where each step the linearization is known. 
\begin{align}
    \delta x(\bar{t}_{j}) &\approx f_{x,\Delta_j}\delta x(t_{k}) + f_{u,\Delta_j} \delta u(t_{k})\\
    \delta x(\underbar{$t$}_{j+1}) &\approx \Xi\delta x(\bar{t}_{j})\label{eq:saltation_expansion}
\end{align}
where $f_{*,\Delta_{j}} = D_*f_{\Delta_j}(x,u)$ and the saltation matrix is abbreviated as $\Xi = \Xi_{(I_{j},I_{j+1})}(\bar{t}_{j},x(\bar{t}_{j}),u(t_k))$

These linearization steps can be combined and directly substituted in the coefficient expansion equations \eqref{eq:expansion_x}--\eqref{eq:expansion_uu} in place of the $f_k$ terms. For the transition cost, $J_{N_j}$, an expansion is taken about $\delta x(\bar{t}_{j})$ which can be mapped back to $(\delta x(t_k),\delta u(t_k))$ and added to the expansion coefficients. When combining all the expansion terms, the hybrid iLQR coefficients in \eqref{eq:delta_Q} are,
\begin{align}
    Q_{x,k}  &=  J_{x}+f_{x,\Delta_{j}}^T J_{x,N_j} + f_{x,\Delta_j}^T\Xi^T V_{x} \label{eq:transition_expansion_first}\\
    Q_{u,k}  &=  J_{u} +f_{u,\Delta_{j}}^T J_{x,N_j} + f_{u,\Delta_{j}}^T\Xi^T V_{x}\\
    Q_{xx,k} &= J_{xx}+f_{x,\Delta_{j}}^TJ_{xx,N_j} f_{x,\Delta_{j}}+ f_{x,\Delta_j}^T\Xi^T V_{xx} \Xi f_{x,\Delta_j}\\
    Q_{ux,k} &= J_{ux}+f_{u,\Delta_{j}}^TJ_{xx,N_j} f_{x,\Delta_{j}} + f_{u,\Delta_j}^T\Xi^T V_{xx} \Xi f_{x,\Delta_j}\\
    Q_{uu,k} &= J_{uu}+f_{u,\Delta_{j}}^TJ_{xx,N_j} f_{u,\Delta_{j}}+ f_{u,\Delta_j}^T\Xi^T V_{xx} \Xi f_{u,\Delta_j}\label{eq:transition_expansion}
\end{align}
After this update to the coefficient expansion, the backwards pass continues normally. 
If the second order variational expression for the saltation matrix is calculated, then these exact changes can be used for a hybrid DDP version of this backwards pass.
However, the computation of the second order variation expression may not be easy for systems with large state space.
Note that a simplification we make for the expansion is that
we assume that the hybrid transition occurs at the end of
the timestep to ensure piecewise smooth control inputs.
This simplification will no longer be a good approximation as time step periods become longer.

\subsection{Hybrid extensions for mode mismatches}\label{section:hybrid_extensions}
Since the forward pass can alter the contact sequence, the new trajectory is not confined to the previous trajectory's mode sequence or timing. This feature is intended because the algorithm can now remove, add, or shift mode transitions if cost is reduced. However, this introduces an issue when the reference mode is not the same as the current mode. 

In \cite[Eq.~7]{rijnen2015optimal}, the authors consider the problem of mode mismatch for an optimal hybrid trajectory, both of the reference and of the feedback gains -- the reference is extended by integration, and the gains are held constant.
We employ their strategy, as well  as apply this same rule for the feedforward input and the feedforward gains -- applying the input intended for a different mode can cause destructive results, or be not well-defined.
If the number of hybrid transitions exceeds that of the reference, we elected to hold the terminal state and gains constant, though other choices could be made instead.

\subsection{Algorithm} \label{section:algorithm}

With each hybrid modification to iLQR listed in Sections \ref{section:hybrid_forwards}, \ref{section:hybrid_backwards}, and \ref{section:hybrid_extensions} our new algorithm can be summarized as follows: 1) Given some initial state, input sequence, quadratic loss function, number of timesteps, and timestep duration a rollout is simulated (either through event driven or time stepping methods) to get the initial reference trajectory and mode sequence. 2) A hybrid backwards pass (using the regularization from \cite{mayne1973differential}) computes the optimal control inputs for the reference trajectory. 3) Hybrid reference extensions are computed on the start and end states for each hybrid reference segment. 4) The forward pass simulates the current mode's dynamics until a hybrid guard condition is met or it is the end of the simulation time; if the guard is reached, the corresponding reset map is applied and the simulation is continued. This forward pass is repeated with a different learning rate until the line search conditions are met \cite{tassa2012synthesis}. 5) Then the backwards pass, hybrid extensions, and forward passes are repeated until convergence.

\section{HiLQR MPC Implementation}
\label{sec:HiLQR_MPC}
In this section, the tracking problem is defined, and we show how to adapt Hybrid iLQR to be a MPC controller.

\subsection{Hybrid Cost Update}
\label{subsec:hybridcost}
{HiLQR for trajectory optimization's goal is to reach a specific state whereas HiLQR MPC is now trying to reach every point a long a trajectory at specific points in time.} The goal {now becomes minimizing} the difference in state and input with respect to a reference state and input
\begin{align}
    {\hat{J}}(x_i,u_i) = (x_i-\hat{x}_i)^T Q_i (x_i-\hat{x}_i) + (u_i-\hat{u}_i)^T R_i (x_i-\hat{u}_i)
    \label{eq:runtime_tracking_cost}
\end{align}
where $Q_i$ is the quadratic penalty matrix on state, and $R_i$ is the quadratic penalty matrix on input, and $(\hat{x},\hat{u})$ denotes the reference.
The optimization problem is now
\begin{align} 
\min_{U} \quad & \hat{J}_N(x_N) + 
      \sum_{i=0}^{N-1} \hat{J}(x_i,u_i) \\ 
\text{where} \quad & x_0 = x(0)\\
& x_{i+1} = f_\Delta(x_i,u_i) \quad \forall i \in \{0, ..., N-1\}
\end{align}

However, because Hybrid iLQR is contact implicit (the hybrid mode sequence can differ from the target's mode sequence), the runtime cost \eqref{eq:runtime_tracking_cost} can be ill defined when the candidate trajectory's mode does not match the target's.
For example, if there is an early or late contact in a rigid body system with unilateral constraints, the velocities will be heavily penalized for having a mismatched timing.
This issue is further propagated to the backward pass, where the gradient information relies on these differences and can ultimately lead to the algorithm not converging.
To mitigate these mode mismatch issues, we propose 2 different solutions for event-driven and timestepping simulations.

For event-driven hybrid simulators, the same hybrid extensions used in reference tracking on the forward pass in Hybrid iLQR can be used when comparing error during mode mismatches. 
Suppose a hybrid transition occurs at time $t$.
The reference state at pre-transition $\hat{x}(t^-)$ is extended beyond the hybrid guard by flowing the pre-transition dynamics forwards while holding the pre-transition input constant. 
The post-transition reference state $\hat{x}(t^+)$ is extended backward by flowing the dynamics backward in time while again holding the input constant.
With these hybrid extensions, when there is a mode mismatch induced by a transition timing error, the reference is switched to the extension with the same hybrid mode.
{Note that when tracking error trends to zero, then the time duration of mode mismatch also trends to zero.
Because of this, the local minimum will not change with references when error goes to zero, and the references will not ultimately change the optimal solution when there is zero error, it only helps to find the optimal solution when error is non zero.}

In timestepping simulations, the effect of the hybrid transition is applied over several timesteps rather than instantaneously as in event-driven hybrid simulations.
For example, when a contact is made, the penetrating velocities do not immediately go to zero and actually take several timesteps to go to zero.
During these timesteps, the hybrid mode is not well defined.
Because of this, the hybrid extension method does not work due to the timesteps that are ``in between'' hybrid modes.
Instead, we propose to use a different approach for legged robots, where the constraint forces $\lambda_j$ are used to scale the penalty on input from $R_{min}$ to $R_{max}$
\begin{align}
    w_j &= \frac{\lambda_j}{\sum_z \lambda_z}\\
    R_j &= R_{max} - w_j(R_{max}-R_{min})
\end{align}
where{ the subscripts} $j${ and $z$} corresponds to the leg index.
{The constraint forces are recorded from the output of Isaac gym simulations and they represent the ground reaction force from each leg of the robot.}
This modification penalizes changes in input less when a leg applies more ground reaction force and penalizes changes in input more when the leg applies less force to the ground. 
This is intuitive because when a leg is not supporting much weight, we want that leg to have lower gains because it has less control authority on the robot body.

\subsection{Rollout and Forward Pass}
\label{subsec:rollout}
\begin{figure}[t]
    \centering
    \includegraphics[width=0.48\textwidth]{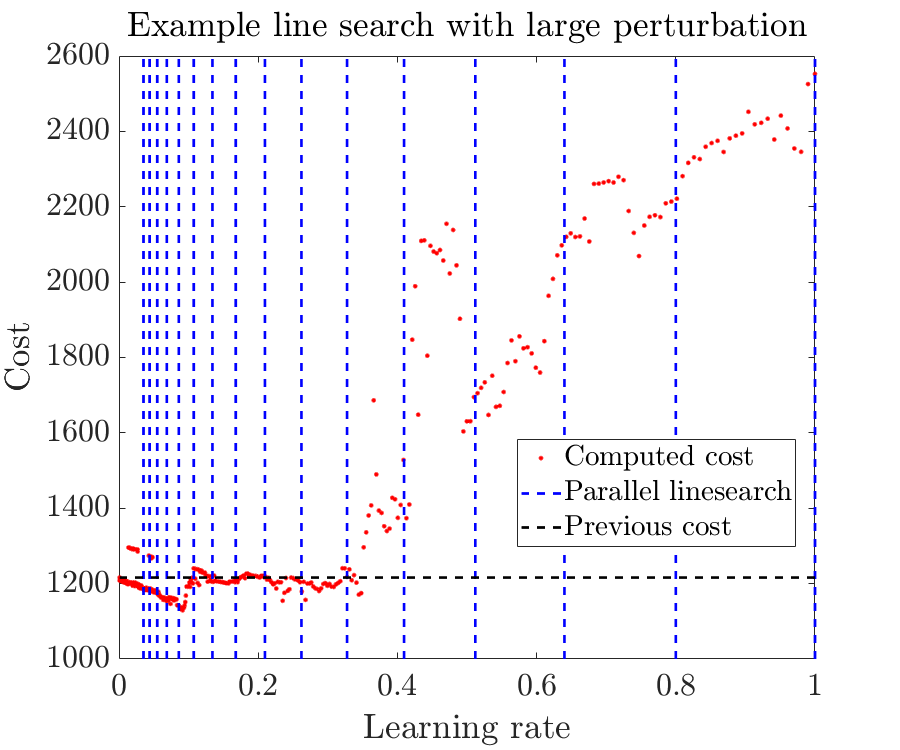}
    \caption{Linesearch for the first HiLQR MPC forward pass iteration after applying a $1.5 \frac{m}{s}$ lateral perturbation while walking as shown in Fig \ref{fig:combined_large_perturbation_first_step}. If computed sequentially, the linesearch would terminate after 12 steps.}
    \label{fig:linesearch}
\end{figure}
Depending on the hybrid system, HiLQR MPC uses either an event-driven or timestepping simulation for its rollouts and forward passes.
In this work, we demonstrate the cost mismatch update for an event-driven simulation on a bouncing ball.
However, when multiple contacts are involved, as in the case for a quadruped robot, simulating an event-driven system is significantly more difficult than using an out-of-the-box timestepping rigid body dynamics simulator.
Many rigid body contact simulators utilize timestepping simulation methods.
In this work, we use ``Isaac Gym'' (a high performance GPU-based physics simulation) \cite{isaacgym}, because the simulator has a unique feature where it can simulate multiple robots at once {at a fraction of the runtime of simulating them serially}.
We utilize parallel computations to parallelize the linesearch in the forward pass.
An example linesearch is shown in Fig. \ref{fig:linesearch}, which shows the cost for different learning rates. 
Note that the cost is discontinuous with respect to the learning rate because the line search explores different contact sequences.
In order for cost to be reduced in this case, the linesearch needs to take 12 steps if done sequentially.
Due to the efficiency of parallel computations on the forward passes, parallelizing is on average twice as fast as computing the linesearch sequentially when comparing the computation times for the solutions in Fig. \ref{fig:combined_large_perturbation_first_step}.
Another approach for ``parallel line search'' \cite{peachey2009parallel} is a method which speeds up optimizations by computing jobs serially, but terminates jobs which take longer to compute.
This method is completely different from our parallel line search method which computes all jobs as once.

Several key implementation features consist of precomputing the gain schedule for the reference trajectory, reusing the valid portions of previous solutions, and always seeding the reference trajectory as one of the parallel solves in the linesearch.

Lastly, quaternion differences \cite{manchester2021planning} are used instead of Euler angles when computing the orientation cost and linear feedback.
This change allows for better convergence properties, as well as allowing for tracking more dynamic behaviors like the backflip in Fig. \ref{fig:backflip} {due to properly accounting for the group structure.
See \cite{manchester2021planning} for more details on quaternion differences and how they improve solving optimization problems when used in place of Euler angles.}

\subsection{Backward Pass}
\label{subsec:backward}
The main challenge for the backward pass is how to compute the derivatives of the dynamics.
For simple hybrid systems like the bouncing ball, the derivatives of the dynamics and saltation matrix are trivial to find and compute.
However, computing the derivatives for the full order rigid body dynamics with unilateral constraints is not trivial -- if done naively, the computations are incredibly slow.
This is the same for the saltation matrix because it relies on computing the derivative of the impact map.
In this work, we utilize a rigid body dynamics library called Pinocchio \cite{pinocchioweb} (which computes these derivatives in an optimized fashion) for all full order contact rigid body dynamics derivatives.
{This work adds onto using Pinocchio's analytical derivative computation capabilities by efficiently computing the Saltation matrix which provide analytical.
Again, the saltation matrix is applied when an MPC horizon adds a new contact.}

For the backward pass, HiLQR MPC assumes the trajectory is produced by an event-driven simulation.
If the timesteps are small enough, then approximating a timestepping simulation as an event-driven simulation on the backward pass is reasonable.
Another approximation HiLQR MPC makes is that when simultaneous contacts are made during a timestep 
{\cite{rijnen2019sensitivity}} (i.e., 2 feet making contact at the same time), the contact sequencing is assumed to always follow the same contact order and to have happened at the end of the timestep.
The chosen order is in increasing order of the indexing of the limbs.
{This approximation effectively chooses a single contact sequence for the resulting Bouligand derivative \cite{burden2016event} instead of using the entire tree of possible contact sequences.
However, choosing an arbitrary contact sequence during a simultaneous impact event is a good approximation for our class of system -- see \cite{council2022representing} for more details.}
These approximations are validated through experimentation, where HiLQR MPC is still able to converge with these approximations in the presence of perturbations.

\subsection{General Robot Implementation}
For all robot experiments using HiLQR MPC, a $50$ timestep MPC horizon is used with timesteps of $0.01$ seconds.
When running HiLQR MPC in simulation, the algorithm is able to pause the simulation in order to compute a new trajectory.
Once a trajectory is generated, the first input of the planned trajectory is used as the control input for that timestep. 
Allowing HiLQR MPC to pause the simulation ensures that we can analyze how well the controller can perform independent of the computation time available.
We also run the controller in real-time, because on hardware the dynamics cannot be paused.
{Note that because perturbations can lead to varying solve times, it is important that the algorithm ran on the robot in real-time have cutoffs on computation time.}

\begin{figure}[t]
    \centering
    \includegraphics[width=0.48\textwidth]{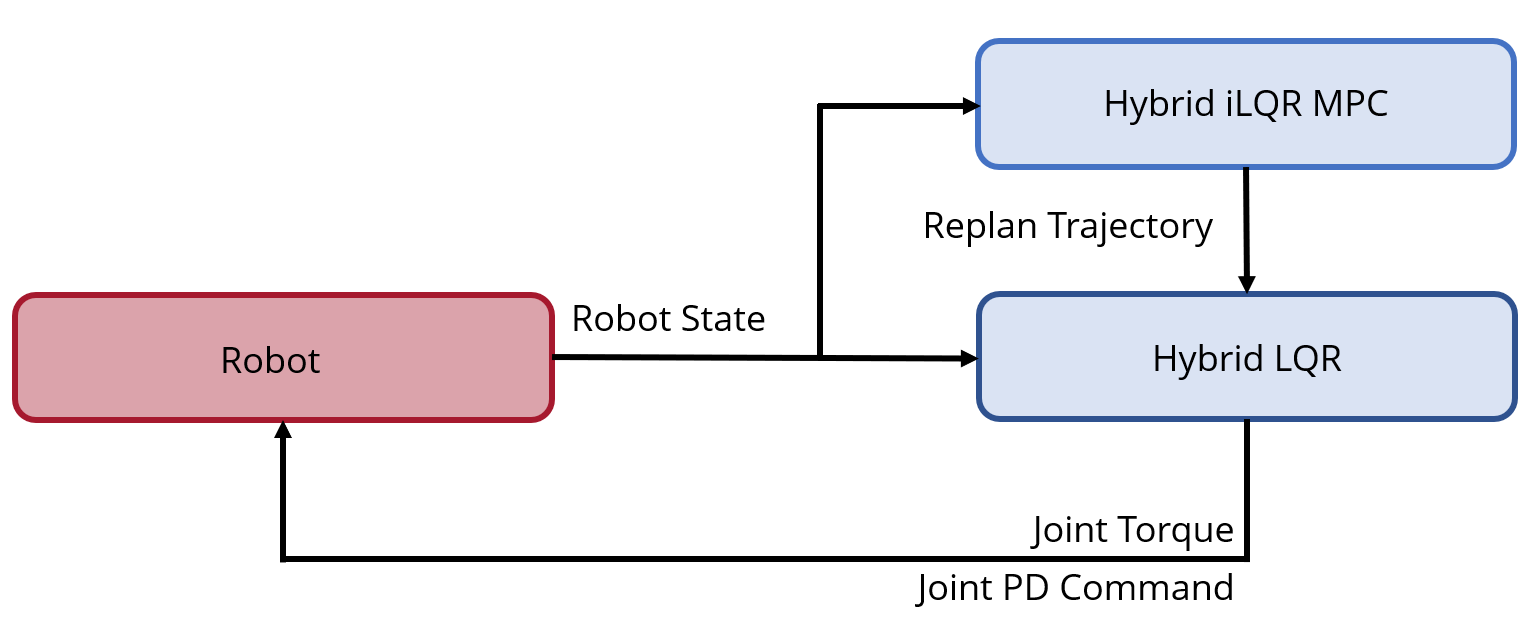}
    \caption{Hierarchy of controllers where HiLQR MPC is replanning trajectories as fast as possible while Hybrid LQR is tracking the most recent trajectory that was sent by HiLQR MPC.}
    \label{fig:ilqr_mpc_diagram}
\end{figure}
To run HiLQR MPC in real-time for the physical robot implementation, several changes are made and hyper parameters are tuned to speed up the algorithm at the cost of performance.
The first change is to run a hierarchy of controllers, as shown in Fig. \ref{fig:ilqr_mpc_diagram}, where a fast low level Hybrid LQR controller is run asynchronously from the trajectory generator (HiLQR MPC).
HiLQR MPC runs separately as fast as possible and always using the latest robot state.
When solving for a new trajectory, sub-optimal trajectories are sent out at each forward pass iteration in order to send the low level controller the most recent trajectory modifications.
If the current solve exceeds the maximum allotted time {0.25s (half of the MPC horizon)}, the current solve is terminated and a new solve is started for the most recent robot state information.
Several hyper-parameters are modified to reduce computation time, from reducing the number of robots running in parallel in the rollouts and forward passes to relaxing the optimality condition. 
Lastly, joint PD terms from the gain matrix are sent directly to the motor controller, which runs at 10KHz rather than computing the feedback at the Hybrid LQR level.
Note that the linearized model is used for tracking, but is constantly updated because the system is inherently nonlinear.

\section{Experiments for HiLQR Trajectory Optimization}
\label{sec:experiment-trajopt}
In this section,
we define a set of hybrid systems -- ranging from a simple 1D bouncing ball to a perching quadcopter with constrained dynamics and friction -- and a series of experiments which evaluates how our hybrid iLQR algorithm performs in a variety of different settings. 
Overall, the Jacobian of the reset map method $D_xR$-iLQR has trouble converging and has worse cost compared to our proposed algorithm $\Xi$-iLQR which uses the saltation matrix.

For all of the examples, we assume that there is no desired reference trajectory to track and that there is no hybrid transition cost $J_{N_j}$ -- this means the runtime cost is only a function of input. 
In each experiment, a comparison against using the Jacobian of the reset map instead of the saltation matrix is made by evaluating the expected cost reduction for the entire trajectory and the final cost. 
The Jacobian of the reset variant is labeled as $D_xR$-iLQR and the main variant which uses the saltation matrix $\Xi$-iLQR. 

For all examples, $m=1$ is the mass of a rigid body, $g=9.8$ is the acceleration due to gravity, the number of timesteps simulated is $N=1000$, and the timestep duration is $\Delta = 0.001$s unless specified.

The dynamics considered here fall into the category of Euler Lagrange dynamics subjected to unilateral holonomic constraints. We use the dynamics, impact law, and complementarity conditions as derived in \cite{johnson2016hybrid}.
These examples use event-driven simulation implemented using MATLAB ODE 45 \cite{shampine2003solving} with event detection.

These systems have configuration variables $q$ where the state of the system is the configurations and their time derivatives $x = [q^T,\dot{q}^T]^T$.
When the system is in contact with a constrained surface $a(q) = 0$, a constraint force $\lambda$ is applied to not allow penetration in the direction of the constraint. The accelerations $\ddot{q}$ and constraint forces $\lambda$ are found by solving the constraint and accelerations simultaneously,
\begin{align}
    M(q)\ddot{q} &+C(q,\dot{q})\dot{q} +N(q,\dot{q})+ A(q)^T\lambda= 
    \Upsilon(q,u)
    \label{eq:constrained_dyn}\\
    A(q)\ddot{q} &+ \dot{A}(q)\dot{q} = 0
\end{align}
where $M(q)$ is the manipulator inertia matrix, $C(q,\dot{q})$ are the Coriolis and centrifugal forces, $N(q,\dot{q})$ are nonlinear forces including gravity and damping, $A(q)  = D_q a(q)$ is the velocity constraint, and $\Upsilon(u)$ is the input mapping function.

Suppose the constrained surface $a_J(q)$ is the $J$th possible hybrid mode, and the current mode is the unconstrained mode. $a_J(q)$ acts as the guard surface for impacts  $g_{(1,J)} = a_J(q)$. When the system hits the impact guard, the velocity is reset using a plastic or elastic impact law \cite{johnson2016hybrid}.

Releasing a constrained mode (liftoff) occurs when a constraint force becomes attractive rather than repulsive; thus we define hybrid guard $g(t,x,u) := \lambda$ and the reset map at these events are identity transforms because no additional constraints are being added.

\subsection{Bouncing ball elastic impact}
\label{sec:bouncing_ball}
\subsubsection{Experimental Setup}
We begin with a 1D bouncing ball under elastic impact \cite{goebel2009hybrid}, where the state $x=[z,\dot{z}]^T$ is the vertical position $z$ and velocity $\dot{z}$. 
The input $u$ is a force applied directly to the ball.
The two hybrid modes, $1$ and $2$, are defined when the ball has negative velocity $\dot{z}<0$ and when the ball has non-negative velocity $\dot{z}\geq 0$, respectively.
The dynamics on each mode are ballistic dynamics plus the input
\begin{align}
    F_1(x,u) = F_2(x,u) := \left[\dot{z},\frac{u-mg}{m}\right]^T
\end{align}
Hybrid mode $1$ transitions to $2$ when the ball hits the ground, $g_{(1,2)}(x) := z $, and mode $2$ transitions to $1$ at apex $g_{(2,1)}(x) := \dot{z}$.
When mode $1$ transitions to $2$, an elastic impact is applied, $R_{(1,2)}(x) = [z,-e \dot{z}]^T$ where $e$ is the coefficient of restitution.
The reset map from $2$ to $1$ is identity.

The Jacobian of the reset map and saltation matrix are,
\begin{align}
    D_xR_{(1,2)} = \begin{bmatrix}
    1 & 0\\
    0 & -e
    \end{bmatrix}, \, \Xi_{(1,2)} = \begin{bmatrix}
    -e & 0\\
    \frac{(u-mg)(e+1)}{m\dot{z}} & -e
    \end{bmatrix}
\end{align}
When transitioning from $2$ to $1$, both Jacobian of the reset map and saltation matrix are identity.

The problem data is to have the ball fall from an initial height with no velocity, $x_0 = [4,0]^T$, and end up at a final height $x_{des}$ with no velocity with penalties $R = 5\times10^{-7}/\Delta$, $Q_N = 100I_{2\times 2}$ and the problems were seeded with a constant input force to obtain different number of bounces.
A suite of bouncing conditions are considered and are summarized in Table \ref{table::experiment_table}. 
In the case where $0$ bounces are optimal $x_{des} = [3,0]^T$ while
where $1$ or $3$ bounces are optimal $x_{des} = [1,0]^T$. For $3$ bounces the timestep is set to $\Delta = 0.004$.
To evaluate the effectiveness of the hybrid extensions, Sec.~\ref{section:hybrid_extensions}, an additional comparison using our hybrid iLQR algorithm where we do not apply any hybrid extensions is made. 
For all cases, a convergence cutoff for this problem is set to be if $|\delta J|\leq 0.05$.

\begin{table}[tb]
\vspace{1em}
\caption{{Bouncing ball with elastic impacts. Trials vary in optimal number of bounces, number of seeded bounces, which method was used, total cost, and convergence $|\delta J| < 0.05$}}

\centering
\begin{tabular}{c c c c c c}
\hline
Optimal $\#$ & Seed $\#$ & Method & Actual $\#$& Cost  & Converged\\
\hline
0&0& $\Xi$ & 0 & $53.1$ & Yes\\
0&0& $D_xR$ & 0 & $53.1$ & Yes\\
0&1& $\Xi$ & 1 & $114$ & Yes\\
0&1& $D_xR$ & 0 & $53.1$ & Yes\\
0&1& Direct & 1 & $114$ & Yes\\
\hline
1&0& $\Xi$ & 0 & $97.3$ & Yes\\
1&0& $D_xR$ & 0 & $97.3$ & Yes\\
1&1& $\Xi$ & 1 & $42.5$ & Yes\\
1&1& $D_xR$ & 0 & $97.3$ & Yes\\
1&3& $\Xi$ & 1 & $42.5$ & Yes\\
1&3& $D_xR$ & 1 & $125$ & No\\
\hline
3&1& $\Xi$ & 1 & $105$ & Yes\\
3&1& $D_xR$ & 0 & $114$ & Yes\\
3&3& $\Xi$ & 3 & $0.536$ & Yes\\
3&3& $D_xR$ & 3 & $19.6$ & No\\
3&3& No Ext. & 3 & $53.3$ & No\\
\hline
\label{table::experiment_table}
\end{tabular}
\end{table}

\subsubsection{Results}
\begin{figure}[t]
    \centering
    \vspace{0.5em}
    \includegraphics[width = \columnwidth]{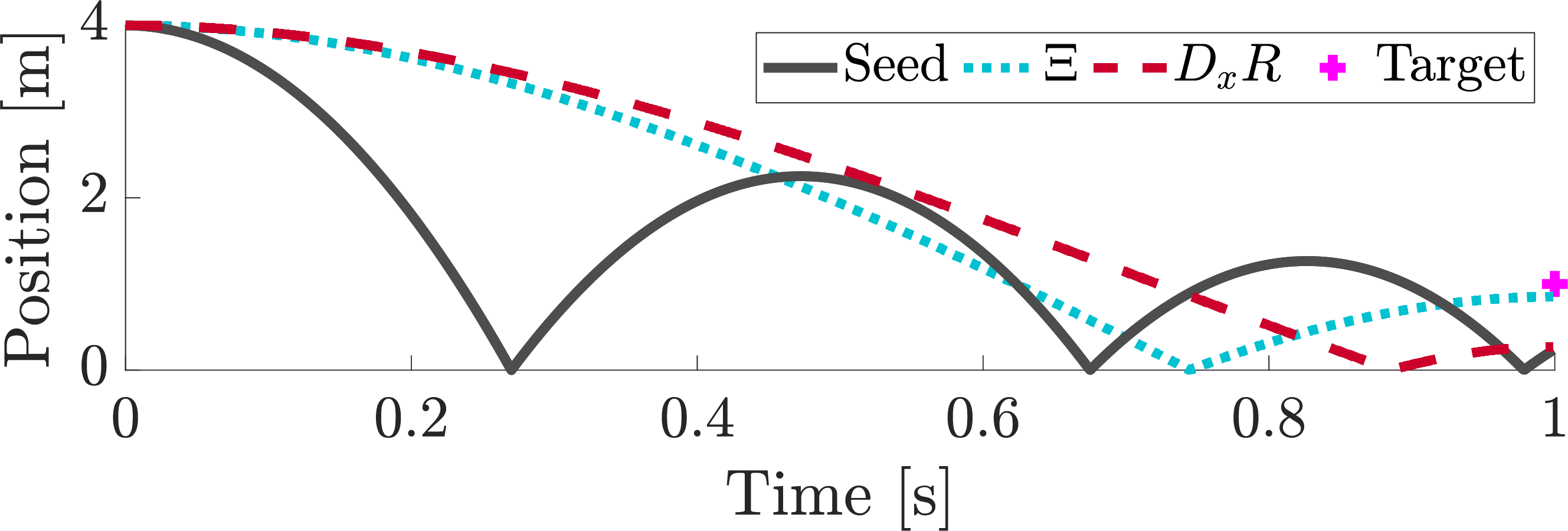}
    \caption{{Bouncing ball with elastic impact where 1 bounce is optimal and 3 bounces are seeded. The target end position is shown in (magenta plus).
    Both gradient update methods were able to pull away the unnecessary bounces, but the method using $D_xR$ did not converge or get to the target state.}}
    \label{fig:triple_bounce}
\end{figure}
The outcomes of the experiment comparing $D_xR$-iLQR to $\Xi$-iLQR are shown in Table \ref{table::experiment_table}. An example run is shown in Fig. \ref{fig:triple_bounce}.
$D_xR$-iLQR did not converge ($|\delta J|>0.05$) on any example if a hybrid transition was maintained, while $\Xi$-iLQR converged on every example. 
The only cases where $D_xR$-iLQR converged were when the algorithm removed all of the bounces -- 
which becomes equivalent to smooth iLQR.
$\Xi$-iLQR has lower cost compared to $D_xR$-iLQR for every example except for when the problem is seeded with no bounces (they obtain the same smooth solution) and when no bounces was the optimal solution but the problem was seeded with a single bounce -- which is an antagonistic seed for the problem. 
In this case, $\Xi$-iLQR did converge to a different local minima\footnote{
This solution was confirmed as a local minima under a single bounce by comparing it against a trajectory produced using direct collocation \cite{kelly2017introduction} constrained to a single bounce, as shown in Table. \ref{table::experiment_table}.
}, which is not surprising as it is not a global optimization. 

The value of the hybrid extension was tested on the three bounce optimal three bounce seeded case. Without the hybrid extension, the optimizer did not converge and did significantly worse than $D_xR$-iLQR. 
This highlights the importance of the hybrid trajectory extensions: even though the backwards pass is correct, having mode mismatches will lead to unfavorable convergence and trajectory quality.

Overall, $\Xi$-iLQR produced locally optimal solutions for each variation and was able to remove unnecessary bounces in some cases, though it never added any. 
This result is expected because there is no gradient information on the backwards pass being provided to give knowledge about adding additional bounces. Furthermore, as discussed above, there may not be an appropriate controller available when a novel hybrid mode is encountered.

\subsection{Ball dropping on a spring-damper}
\subsubsection{Experimental Setup}
Hard contacts are sometimes relaxed using springs and dampers, so we consider the 1D bouncing ball case, but instead of having a discontinuous event at impact, the impact event is extended by assuming the ground is a spring damper (i.e., a force law  $f_{sd}(z,\dot{z}) := kz+d\dot{z})$ when being penetrated and a spring when releasing.
The system now has an identity reset, but since the saltation matrix is not identity, the hybrid transition still produces a jump in the linearization.

The hybrid modes are defined as: the aerial phase $1$, the spring-damper phase $2$ and the spring phase $3$.
The spring and dampening coefficients are chosen to be $k = 100$ and $d = 5$.
The guards are when the ball hits the ground $g_{(1,2)} = z$, when the ball no longer has any penetrating velocity $g_{(2,3)} = \dot{z}$, and when the ball is released from the ground $g_{(3,1)} = z$.
For all of these transitions, the reset map is an identity transformation and the states do not change.

The example is setup to have the ball fall an initial height with an initial downwards velocity $x_0 = [3,-2]$, end up at a height with no velocity  $x_{des} = [1,0]$, with penalties $R = 0.0001$, $Q_N = 100 I_{2\times2}$ and no input for the seed.

\subsubsection{Results}
    \begin{figure}[t]
    \centering
    \vspace{0.5em}
    \includegraphics[width = \columnwidth]{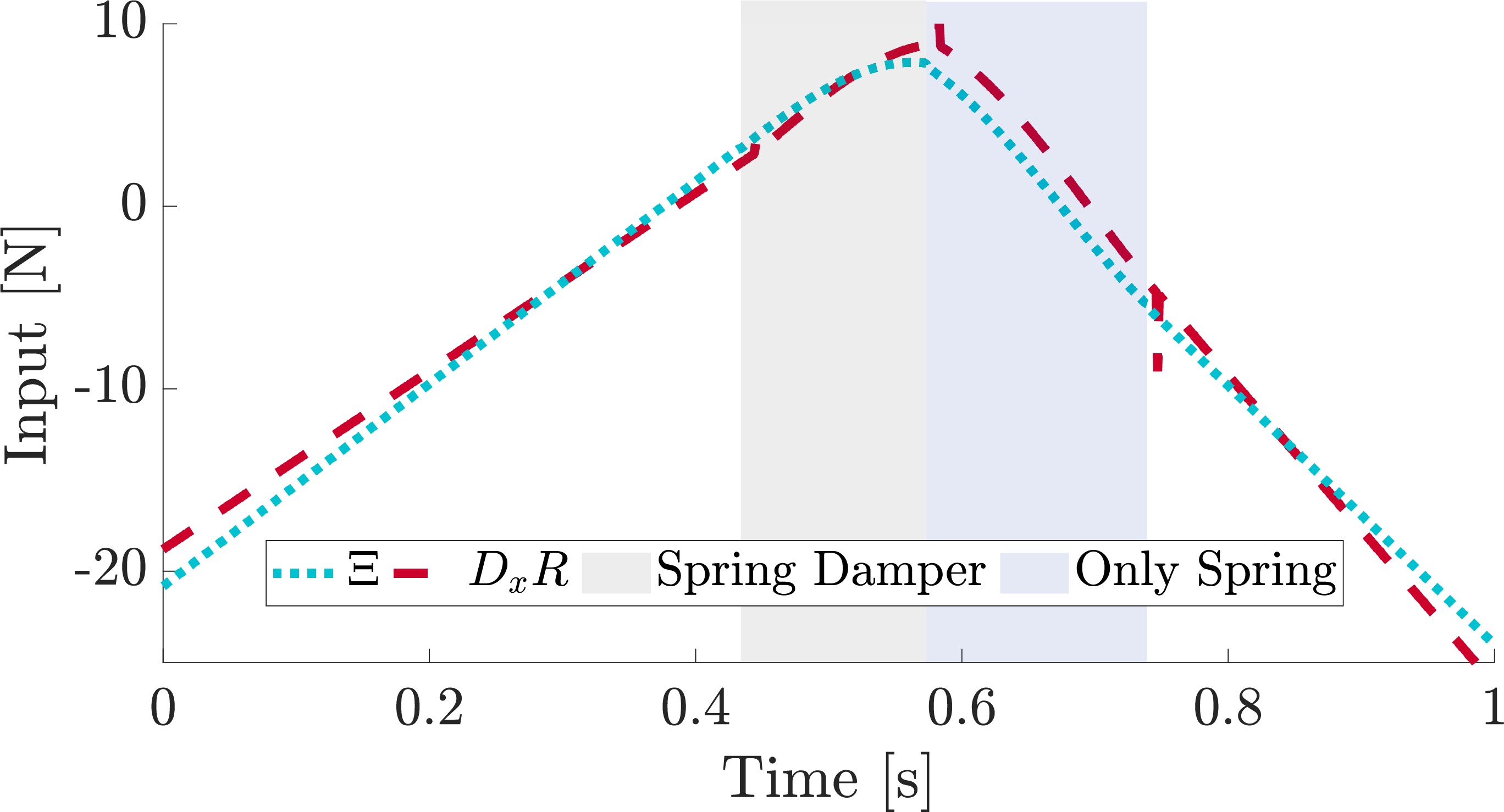}
    \caption{{Bouncing ball on a spring-damper ground where both gradient update methods found similar trajectories but using the Jacobian of the reset map $D_xR$ lead to not being able to fully converge as evident by the residual spikes near hybrid transitions. }
    }
    \label{fig:spring_drop_gains}
\end{figure}
For this experiment, $\Xi$-iLQR and $D_xR$-iLQR came up with similar solutions where the cost of $\Xi$-iLQR $J=13.21$ is slightly lower than $D_xR$-iLQR $J=13.29$. This difference is highlighted in Fig. \ref{fig:spring_drop_gains} where $D_xR$-iLQR was not able to smooth out the spikes near mode changes. 
This is also reflected in $D_xR$-iLQR having a higher expected cost reduction as well $\delta J = 0.001$ where $\Xi$-iLQR is a magnitude lower $\delta J = 0.00017$. 
This difference in convergence can most likely be attributed to $D_xR$ providing gradient information that does not adjust the input pre-impact accordingly to allow for adjustments on the spikes post-impact without destructively changing the resulting end state.

\subsection{Ball drop on a curved surface with plastic impacts}
\subsubsection{Experimental Setup}
To test our algorithm with a nonlinear constraint surface, we designed a system where an actuated ball in 2D space is dropped inside a hollow tube and is tasked to end in a goal location on the tube surface. 

The configuration states of the system are the horizontal  and vertical positions $q = [y, z]^T$.
This system consists of two different hybrid modes: the unconstrained mode $1$ and in the constrained mode $2$. 
The constrained surface is defined to be a circle with radius $2$, $a(q) = 4-y^2-z^2$.
The dynamics of the system, \eqref{eq:constrained_dyn}, are ballistic dynamics with direct inputs on configurations, $M(q) = mI_{2\times2}$, $N(q,\dot{q}) =[0,-mg]^T$, $C(q,\dot{q}) = 0_{2\times2}$, and $\Upsilon=[u_y,u_z]^T$.
The impact guard from (1,2) is defined by the circle's constrained surface and the liftoff guard from (2,1) is the constraint force $\lambda$.

The example is setup to have the ball fall from an initial height with velocity pointing down and to the right $x_0 = [1,0]$, end up at a specific location on circle with no velocity  $x_{des} = [-\sqrt{3},-1,0,0]$, with penalties $R = 0.0001$, $Q_N = 100I_{4\times 4}$ and no input for the initial seed except for a vertical force $2mg$ applied for a small duration to cause the ball to momentarily leave the constraint.

\subsubsection{Results}
    \begin{figure}[t]
    \centering
    \vspace{0.5em}
    \includegraphics[width = \columnwidth]{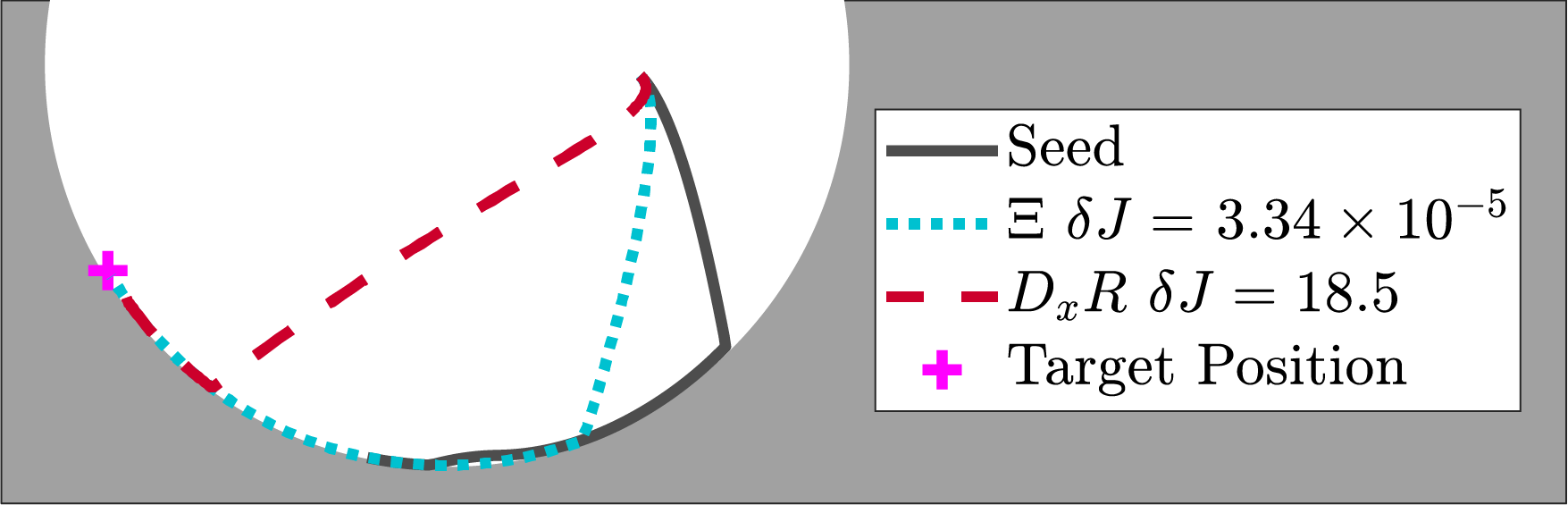}
    \caption{{Ball drop on a curved surface with plastic impacts where both gradient methods produced trajectories that got to the end goal, but using $D_xR$ did not converge and had a significantly higher cost. }
    }
    \label{fig:circle_drop_fig}
\end{figure}
The trajectory produced by $\Xi$-iLQR has a cost of $J=10.7$ and $D_xR$-iLQR a cost of $J=50.5$.
The generated position trajectories along with the initial seeded trajectory are shown in Fig. \ref{fig:circle_drop_fig} where both methods ended up at the goal state but $D_xR$-iLQR converged significantly less than $\Xi$-iLQR.

In this example, we purposely seeded a sub-optimal trajectory which releases the contact for a small duration and returns back to the constraint to evaluate if the algorithms would modify the contact sequence. 
$\Xi$-iLQR ended up removing this erroneous contact change and whereas $D_xR$-iLQR ended up not going back to the constraint surface and ended in the unconstrained mode.
We speculate that because $D_xR$ has the wrong gradient information about contacts, it ended up staying in the unconstrained mode for a longer duration and ultimately could not converge.

\subsection{Perching quadcopter}
\subsubsection{Experimental Setup}
We introduce a quadcopter perching example inspired by \cite{lussier2011landing}, where we consider a planar quadcopter which can make contact with sliding friction on a surface. 
When both edges of the quadcopter are touching the constraint, we assume some latching mechanism engages and fully constrains the quadcopter in place with no way to release.
This problem explores planning with an underactuated system, friction, constraint surfaces, nonlinear dynamics, nonlinear guards, and nonlinear resets.

The configurations of the system are the vertical, horizontal, and angular position $q = [y,z,\theta]^T$ and the inputs are the left and right thrusters, $u_1$ and $u_2$.
The dynamics are defined by \eqref{eq:constrained_dyn} with the following 
\begin{align}
    M(q) &:= \begin{bmatrix}
    m&0&0\\
    0&m&0\\
    0&0&I
    \end{bmatrix}, \quad
    C(q,\dot{q}) := \begin{bmatrix}
    0&0&0\\
    0&0&0\\
    0&0&0
    \end{bmatrix},\\
     N(q,\dot{q}) &:= \begin{bmatrix}
    0\\
    -mg\\
    0
    \end{bmatrix}, \quad
    \Upsilon := \begin{bmatrix}
    -\sin(\theta)(u_1 + u_2)\\
    \cos(\theta)(u_1 + u_2)\\
    \frac{1}{2}(u_2w - u_1w)
    \end{bmatrix}
\end{align}
where $w = 0.25$ is the width and $I = 1$ is the inertia of the quadcopter.

To add more complex geometry, the constrained surface is a circle centered about the origin with radius 5.
Since the edges of the quadcopter make contact with the surface, the left and right edges of the quadcopter are located at,
\begin{align}
    [y_L,z_L]^T &= [y-\frac{1}{2}w\cos{\theta},z-\frac{1}{2}w\sin{\theta}]^T\\
    [y_R,z_R]^T &= [y+\frac{1}{2}w\cos{\theta},z+\frac{1}{2}w\sin{\theta}]^T
\end{align}
The constraints are then $a_1 = 25- y_L^2 - z_L^2$ and $a_2 = 25- y_R^2 - z_R^2$.
Frictional force $\lambda_t$ is defined to be tangential to the constraint with magnitude proportional to the constraint force $\lambda_n$, $\lambda_{t} = \mu \lambda_n$, where $\mu$ is the coefficient of friction. 

The example is setup to have the quadcopter start some distance away from the constraint with a horizontal velocity, $x_0 = [2,2.5,-\pi/8,4,0,0]^T$, end up oriented with the constraint with no velocity $x_{des} = [5\cos(-\pi/12),\allowbreak 5\cos(-\pi/12),\allowbreak -7/12\pi,\allowbreak 0,0,0]^T$, timesteps $\Delta = 0.002$, with penalties $R = 0.01_{2\times2}$, and $Q_N = [1000I_{3\times3},0_{3\times3};\allowbreak  0_{3\times3}, 0.1I_{3\times3}]$.
The position portion is weighted more heavily than velocity because the goal is to get close enough to the desired location to perch.
For the seed, a combined thrust of equal to $1.5mg$ was applied constantly and if both edges made contact with the constraint, the thrust force was dropped to $0.1mg$. This initial input resulted in a trajectory which makes contact with the right edge and then shortly after makes double contact with the constraint as shown in Fig.~\ref{fig:quadcopter_fig}.

\subsubsection{Results}
\begin{figure}[t]
    \centering
    \vspace{0.5em}
    \includegraphics[width = \columnwidth]{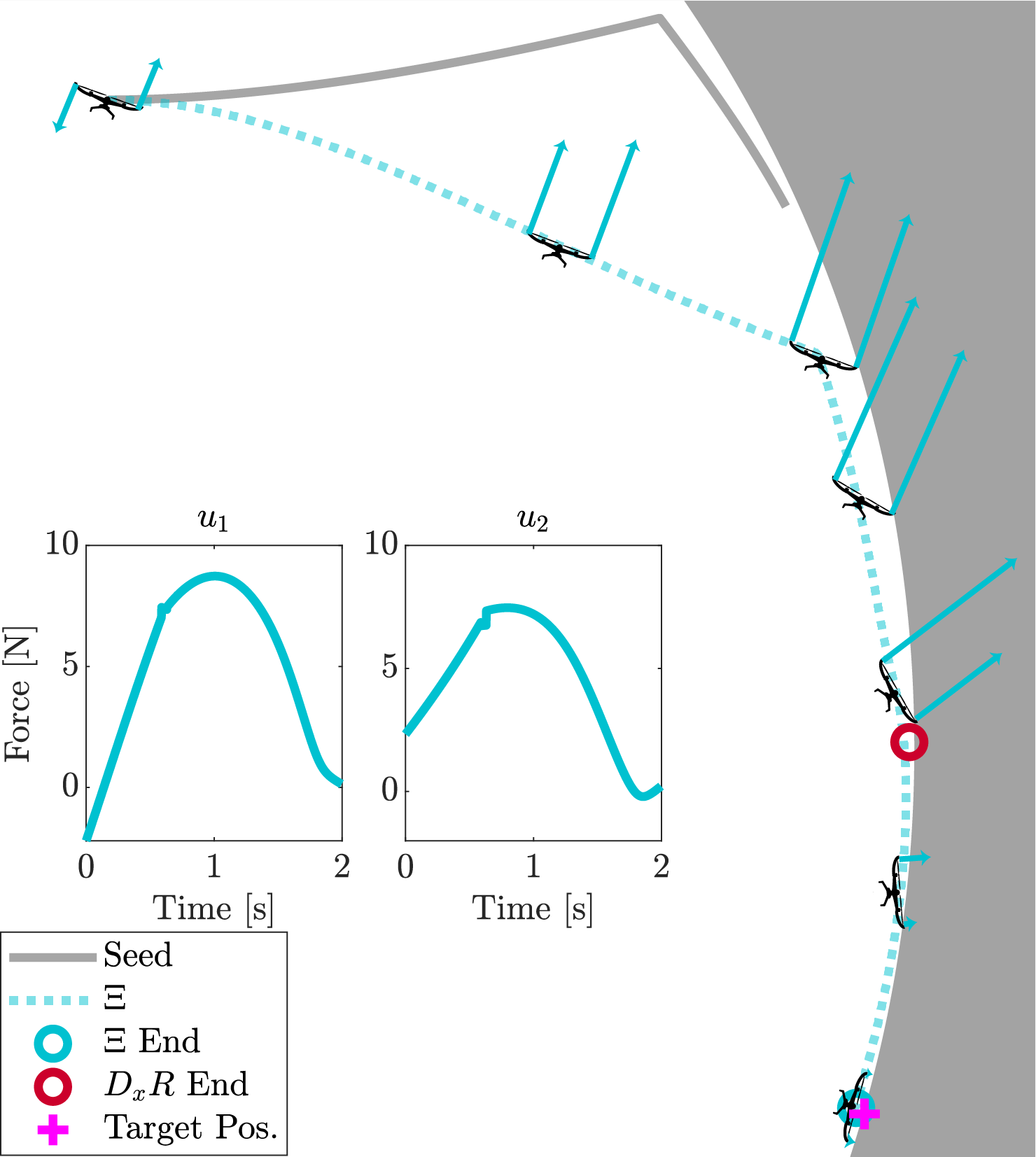}
    \caption{{Demonstrating an example solution using the proposed hybrid iLQR algorithm (labeled with $\Xi$, the saltation matrix, Def.~\ref{def:salt}) where the goal is to control a quadcopter to a target final position (shown with a magenta plus) and can make contact with a curved wall with friction.
    Using a different approximation for the gradient (Jacobian of the reset map, $D_xR$, \cite{li2020hybrid}) leads to poor convergence and significantly higher cost. 
    Note that in the force plots, the optimal input is not smooth because of the hybrid transition.}}
    \label{fig:quadcopter_fig}
\end{figure}

In this example, the final position trajectories are shown in Fig. \ref{fig:quadcopter_fig} where $\Xi$-iLQR converged $\delta J = 0.170$ with a cost of $J = 4.76$ whereas $D_xR$-iLQR did not converge $\delta J = 3\times10^5$ and produced an erratic solution with very high cost of $J = 2.66\times10^3$. 

$\Xi$-iLQR seemed to make the natural extension of the seed and followed the constraint until the target position was achieved, but removed the double constrained mode at the end. 
We postulate that the fully constrained mode was removed in order to better fine tune the final position because position error is weighted significantly more than velocity.
However, the true optimal solution should include the fully constrained mode to eliminate any velocity for free.

\section{Experiments for HiLQR MPC}
\label{sec:mpc_experiments}

In this section, the {experiments and results} for HiLQR {used as an} MPC are presented.
{Overall, we find that utilizing the cost mismatch updates is crucial for obtaining good solutions, and HiLQR MPC can withstand large perturbations by modifying the contact sequence in an optimal manner.
Experimental results are also shown in the video attachment.}

To validate the event-driven mode mismatch cost update, we first compare using the proposed update with not using any hybrid cost updates on a simple actuated bouncing ball hybrid system.
Then, to show how this approach can scale up to a real system, simulated and physical robot experiments are carried out on a quadrupedal robot (Unitree A1) to compare HiLQR MPC with methods that use centroidal simplifications and Raibert heuristics for swing leg control: ``Convex MPC'' \cite{di2018dynamic} and ``Instant QP'' \cite{da2020learning,xie2021glide,gehring2013control}.
Convex MPC returns ground reaction forces for the feet that are in contact with the ground and are subjected to friction constraints for a set horizon length.
The dynamic model is a linearized floating base model and the optimization is formulated as a quadratic program.
Instant QP solves the same problem, but for a single timestep.
Because only one timestep is solved, Instant QP can update the solver with the actual contact condition of the feet and can provide more stability with respect to contact mismatches, but lacks the robustness that is gained from looking ahead. 

\subsection{Bouncing Ball}
\label{subsec:bouncing_ball_MPC}
\subsubsection{{Experimental Setup}}
In this experiment, the same 1D bouncing ball hybrid system from {Sec.~\ref{sec:bouncing_ball}} is used.
To validate that updating the cost on mode mismatches improves convergence for HiLQR MPC, we first generate a reference trajectory using Hybrid iLQR to create an optimal single bounce trajectory.
HiLQR MPC is used to stabilize an initial large perturbation and is run with and without the hybrid cost update for event-driven simulations.
For both cases, HiLQR MPC is applied at every timestep.
At each timestep, convergence is recorded where convergence is determined by the expected reduction \eqref{eq:expected_reduction}.
For this test, the convergence cut-off is set to be $\delta J < 1e^{-4}$.
It is expected that, by utilizing the mode extensions, convergence will improve because the algorithm will not spend unnecessary computation and effort in flipping the velocity of the ball if there is a mismatch in impact timing, rather it will wait for when the impact applies the flip.

\subsubsection{{Results}}
\begin{figure*}[htb]
\centering
\includegraphics[width=\textwidth]{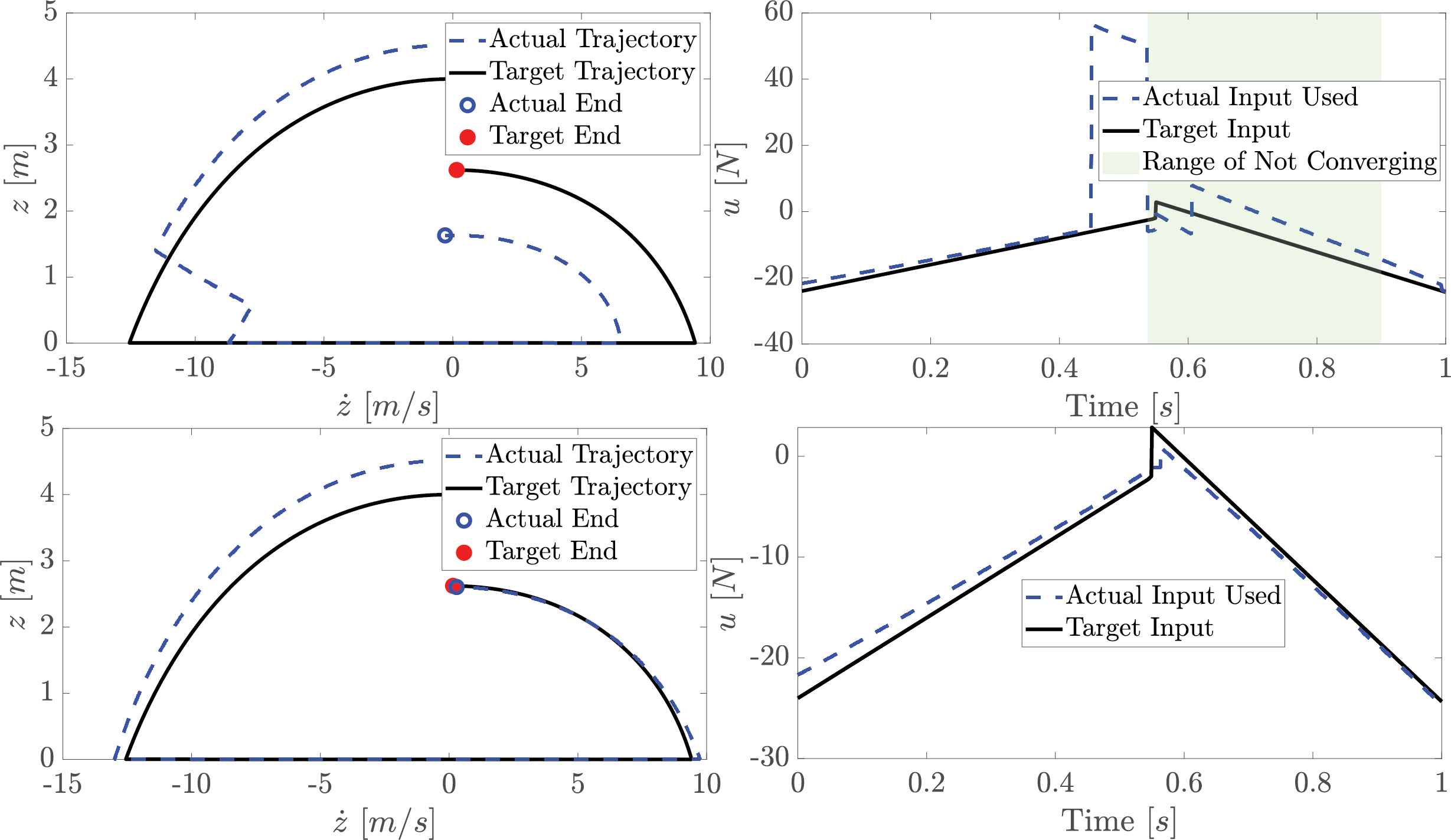}
\caption{Comparing HiLQR MPC not using the event-driven hybrid cost update (top row) and using the event-driven hybrid cost update (bottom row) where the state space trajectory tracking is shown in (left column) and input usage in time series is shown in (right column).
HiLQR solutions are shown in (blue dashed) and the target trajectory is shown in (black solid).
The end of trajectories are denoted with (circle).
When not using the event-driven hybrid cost update the trajectory tracking suffered, as evident by the high input effort and sharp deviations in trajectory that attempt to track the post-impact velocity before the impact occurs.  Several solutions did not converge as shown in with (green highlight).
Whereas, using the event-driven hybrid cost update led to altogether better convergence and tracking.}
\label{fig:bouncing_ball_mpc}
\end{figure*}

The task for the bouncing ball experiment is to track a predefined trajectory using HiLQR MPC for a fully actuated bouncing ball.
The target trajectory is 1 second long, where the ball starts at 4 meters above the ground with no velocity and ends at 2.5 meters above the ground with no velocity.
We compare using the event-driven hybrid cost update (Sec.~\ref{subsec:hybridcost}) to not using this update, and the results of this experiment are shown in Fig. \ref{fig:bouncing_ball_mpc}.

As expected, both methods converge and track well before the impact event is within the horizon of the HiLQR MPC.
The approaches differ once the hybrid event is within the horizon, as can be seen by the high control effort and unnatural kink in state space that is produced when not using the cost update.
Furthermore, of the $1001$ time steps, $8$ did not converge when the cost update was not used.
Although the number of unconverged timesteps is small, the quality of the trajectory suffered greatly, as shown in Fig. \ref{fig:bouncing_ball_mpc}, top row.
This is because without updating the cost to account for hybrid mode mismatches, the gradient information biases the solution towards flipping the velocity before impact.

Using the cost update for hybrid mode mismatches, HiLQR MPC can correctly utilize the impact to reduce tracking error, as shown in Fig. \ref{fig:bouncing_ball_mpc}, bottom row.
The cost update allows HiLQR MPC to create plans that are closer to the target trajectory by shifting contact times rather than making large modifications to the input to match the contact schedule, which results in significantly better convergence.
In addition to having better tracking performance, when using trajectory optimization for MPC, it is desirable to always converge and to not make drastic changes from the planned trajectory unless necessary.



\subsection{Simulated Robot Controller Comparison}
\label{sec:simulated}
\subsubsection{{Experimental Setup}}
To demonstrate the robustness of cohesively planning whole body motions and allowing contact schedules to change, we compare HiLQR MPC to Convex MPC and Instant QP by applying perturbations to A1 while implementing a walking gait in simulation.
To make the comparison fair, the walking gait that HiLQR MPC is tracking is the same one generated from Convex MPC in the absence of perturbations.
Similar gait parameters are chosen for Instant QP to produce a similar gait. 
All controllers are run at each timestep and use the first control input of the new trajectory as the control input for that timestep.

The walking gait starts from a standing pose and then attempts to reach a desired forward velocity of $0.2\frac{m}{s}$.
Lateral velocity perturbations are applied to the robot's body at two different magnitudes and eight different times along the gait cycle: four when each foot is in swing when getting up to speed and the other four when the gait is in steady state.
The number of times the robot falls and the maximum perturbed lateral position are recorded for each push.

It is expected that HiLQR MPC should be able to recover from a wider variety of perturbations and have less deviation when the perturbations are large when compared against the centroidal methods because it can utilize the nonlinear contact dynamics of the swing and stance legs cohesively.

\subsubsection{{Results}}
\begin{figure}[t]
    \centering
    \includegraphics[width=0.45\textwidth]{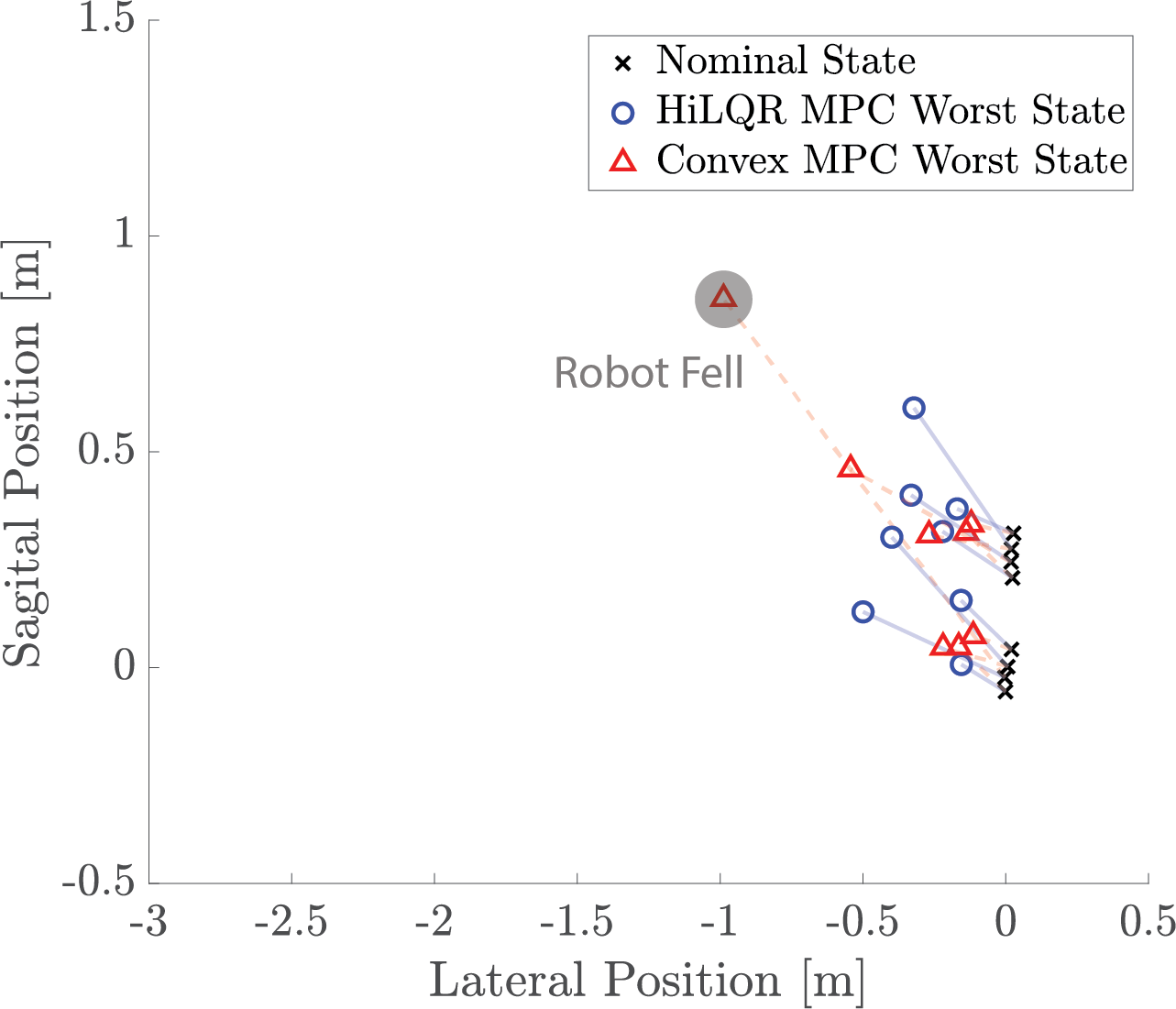}
    \caption{Medium perturbation (1.0 m/s lateral perturbation). Plots the nominal trajectory and worst case error in lateral position for both controllers. 
    {Grey circle indicates that the robot fell.}}
    \label{fig:medium_pert_comparison}
\end{figure}

\begin{figure}[t]
    \centering
    \includegraphics[width=0.45\textwidth]{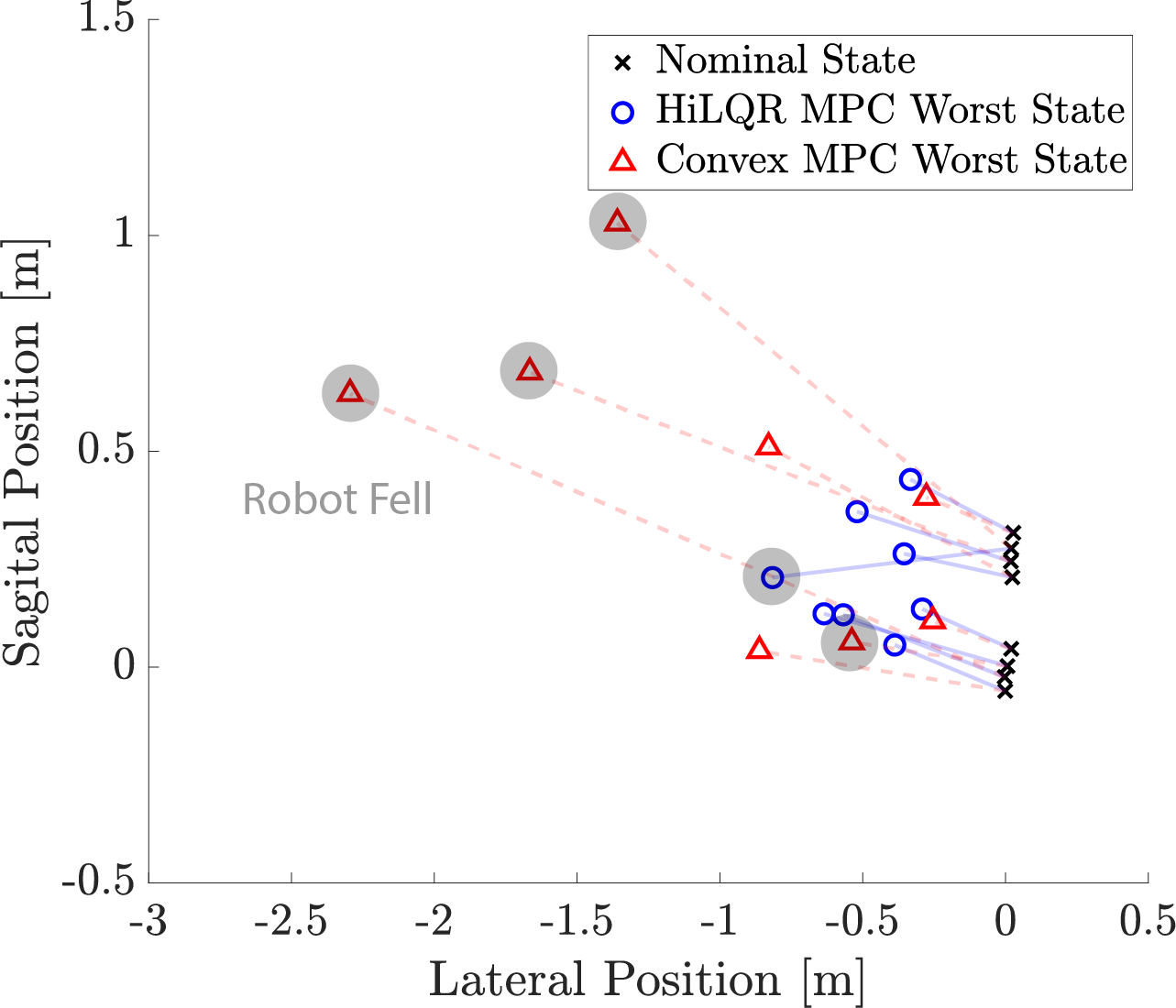}
    \caption{Large perturbation (1.5 m/s lateral perturbation). Plots the nominal trajectory and worst case error in lateral position for both controllers.
    {Grey circle indicates that the robot fell.}}
    \label{fig:large_pert_comparison}
\end{figure}

\begin{figure}[t]
    \centering
    \includegraphics[width=0.45\textwidth]{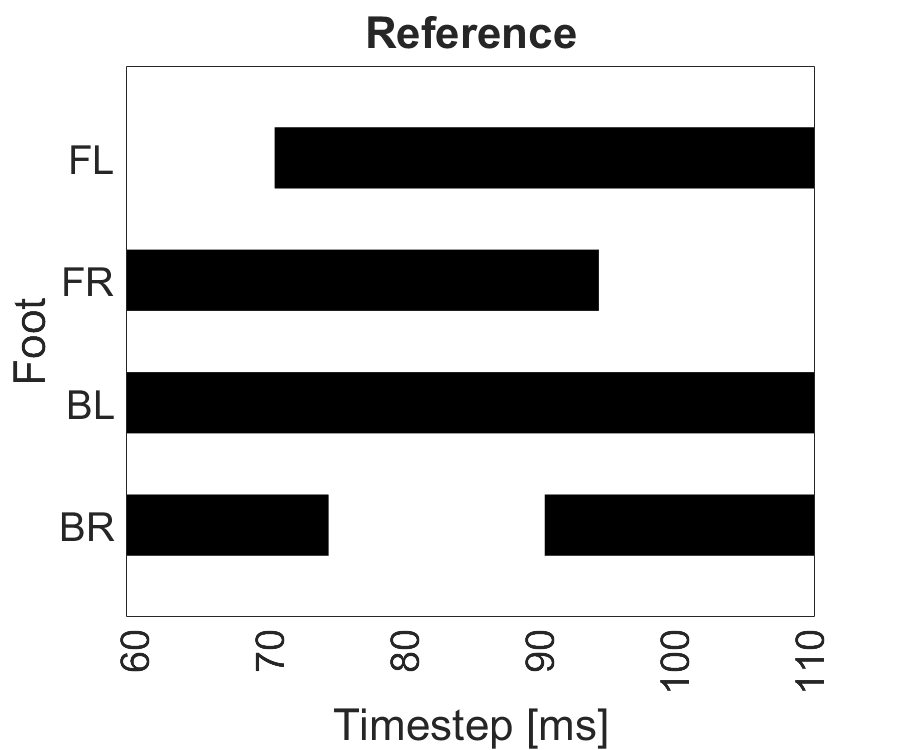}
    \caption{Hildebrand diagram for the nominal walking gait where black means the foot is in contact.}
    \label{fig:hildebrand_nominal}
\end{figure}

\begin{figure}[t]
    \centering
    \includegraphics[width=0.45\textwidth]{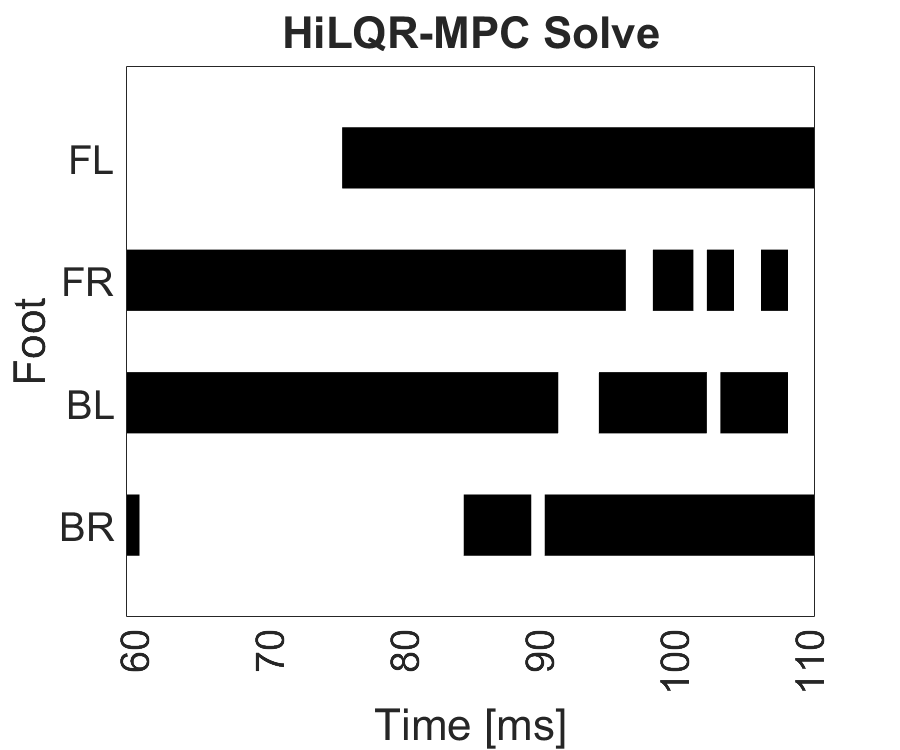}
    \caption{Hildebrand diagram for a single solve of HiLQR MPC rejecting the large perturbation at $60$ ms as shown in the top of Fig. \ref{fig:combined_large_perturbation_first_step} where black means the foot is in contact. See that HiLQR MPC is removing and adding back contacts when advantageous to help stabilize the behavior.}
    \label{fig:hildebrand_ilqr}
\end{figure}

\begin{figure*}[t]
    \centering
    \includegraphics[width=0.95\textwidth]{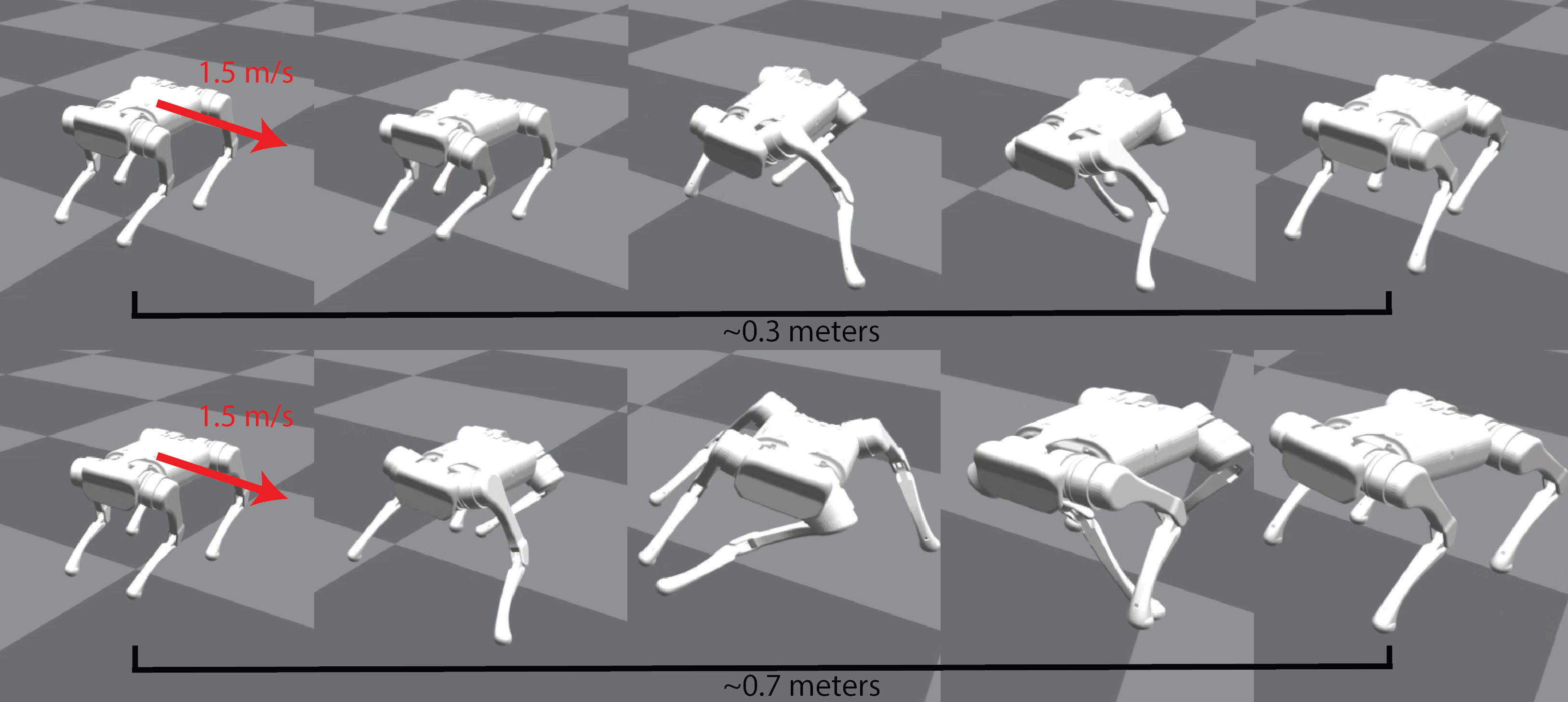}
    \caption{Applying 1.5 m/s lateral perturbation during the first step of the walking gait. (Top row) shows HiLQR MPC recovering from the perturbation in one step and accruing a lateral deviation of 0.3 meters while (bottom row) shows Convex MPC taking several steps to handle the perturbation and is perturbed 0.7 meters away from the nominal.}
    \label{fig:combined_large_perturbation_first_step}
\end{figure*}

The robustness of HiLQR MPC is compared with Convex MPC and Instant QP for a walking trajectory at eight different perturbations in simulation.
The results are summarized in Table \ref{table::lateral_pert}, farthest perturbed position is visualized for each experiment in Figs. \ref{fig:medium_pert_comparison} and \ref{fig:large_pert_comparison}, change in contact sequence in Figs.~\ref{fig:hildebrand_nominal} and \ref{fig:hildebrand_ilqr}, and the resuling behavior shown in Fig~\ref{fig:combined_large_perturbation_first_step}.

As expected, deviations from the smaller perturbation lead to similar results and high success for all controllers.
This is most likely because the perturbations do not require the controller to heavily modify the trajectory while stabilizing less stable robot states, such as in the case of the larger perturbations.
In the medium and large perturbation experiments, HiLQR MPC had a higher success rate of 100\% and 88\% compared to the centroidal methods -- Convex MPC 88\% and 50\% and Instant QP 50\% and 25\%.
Failure for the controllers tended to occur when a right leg was in swing (both front right and back).
This failure mode is most likely due to the lateral perturbation being applied in the left direction causing the stabilizing maneuvers to be more complicated and less stable.
Because HiLQR MPC is able to plan the body and swing legs more cohesively, it can handle these complex maneuvers better than the centroidal methods, where the stance and swing legs are planned separately.
This difference is mostly highlighted when the perturbations are larger.
Since Instant QP performed worse than Convex MPC, further comparisons are made only between HiLQR MPC and Convex MPC.

\begin{table}[tb]
\setlength{\tabcolsep}{3pt}
\caption{Lateral perturbation success rates for a medium perturbation $1.0 {m}/{s}$, a large perturbation $1.5 {m}/{s}$, and the average max deviation for the large perturbation over 8 trials.}
\centering
\vspace{.5em}
\begin{tabular}{l c c c c}
\hline
Controller  & $1.0 {m}/{s}$ Succ.\ [\%] & $1.5 {m}/{s}$ Succ. [\%] & Avg. Dev. [$m$]\\
\hline
HiLQR MPC & $100$\%& $88$\% &$0.512$$m$\\
Convex MPC & $88$\%& $50$\% &$1.032$$m$\\
Instant QP& $50$\%& $25$\% & $3.729$$m$\\
\hline
\label{table::lateral_pert}
\end{tabular}
\end{table}

In the large perturbation experiments, HiLQR MPC deviated half as much as Convex MPC when comparing max lateral deviations in body position, as shown in Table \ref{table::lateral_pert}.
An example trial (large perturbation during the first step) is shown in Fig. \ref{fig:combined_large_perturbation_first_step}, where HiLQR MPC used less steps to stabilize the perturbation, which ultimately led to the body deviating less than half of the deviation from Convex MPC.
The contact sequence for the reference and the initial solution after applying the perturbation are shown in Figures \ref{fig:hildebrand_nominal} and \ref{fig:hildebrand_ilqr}.
Note that HiLQR MPC is solving for new trajectories that modify the contact sequence in order to better stabilize the behavior rather than adhering to the original plan's contact sequence.
This is crucial because HiLQR MPC can add or remove contacts to help catch itself, as well as optimize the new contact locations.

Overall, HiLQR MPC performed similarly to or better than Convex MPC when stabilizing perturbations along a walking trajectory.
When the perturbations are large, HiLQR MPC outperforms Convex MPC because it is able to replan a new contact sequence to stabilize about and it can fully utilize the nonlinear dynamics for the more aggressive maneuvers.

\subsection{Physical Robot Controller Comparison}
\label{sec:physical}
\subsubsection{{Experimental Setup}}
The bulk of the analysis for comparing the controllers is done in simulation because the perturbations can be consistently applied in both cases with a variety of different perturbations.
To reliably apply the same perturbation on hardware, we opt for a consistent motor command block for a short period of time while the robot is walking.
Once the motor commands are unblocked, the controller must react to the robot falling over, catch itself, and then continue walking.

In this experiment, we compare HiLQR MPC against Instant QP.
{Instant QP utilizes similar assumptions to Convex MPC (floating base convex optimization), but differs in being able to update the contact mode on every timestep while Convex MPC follows the contact sequence of the planned gait.
To have a more fair comparison we utilize Instant QP to remove the clear absence of handling a new contact mode that is bound to happen with this experiment.
Both Hybrid iLQR MPC and Instant QP} are able to handle the perturbation in simulation but come up with different solutions.
HiLQR MPC tends to replan a stand trajectory after it realizes that it is falling to catch itself, while Instant QP tries to continue the walking gait and recirculates the legs in order to catch itself.
The perturbation is applied shortly after walking has started, and the torque commands are blocked for $0.15$ seconds.
The experiment is run 5 times for each controller and failure is determined by if the robot's body hits the floor and if the controller is able to continue walking after the perturbation.
For state estimation, we use the Kalman filter from \cite{bledt2018contact}.
Because HiLQR MPC creates a new plan to track in order to handle the perturbation, it is expected to outperform Instant QP which is trying its best to continue walking.


\subsubsection{{Results}}
\begin{figure*}[htb]
    \centering
    \includegraphics[width=0.95\textwidth]{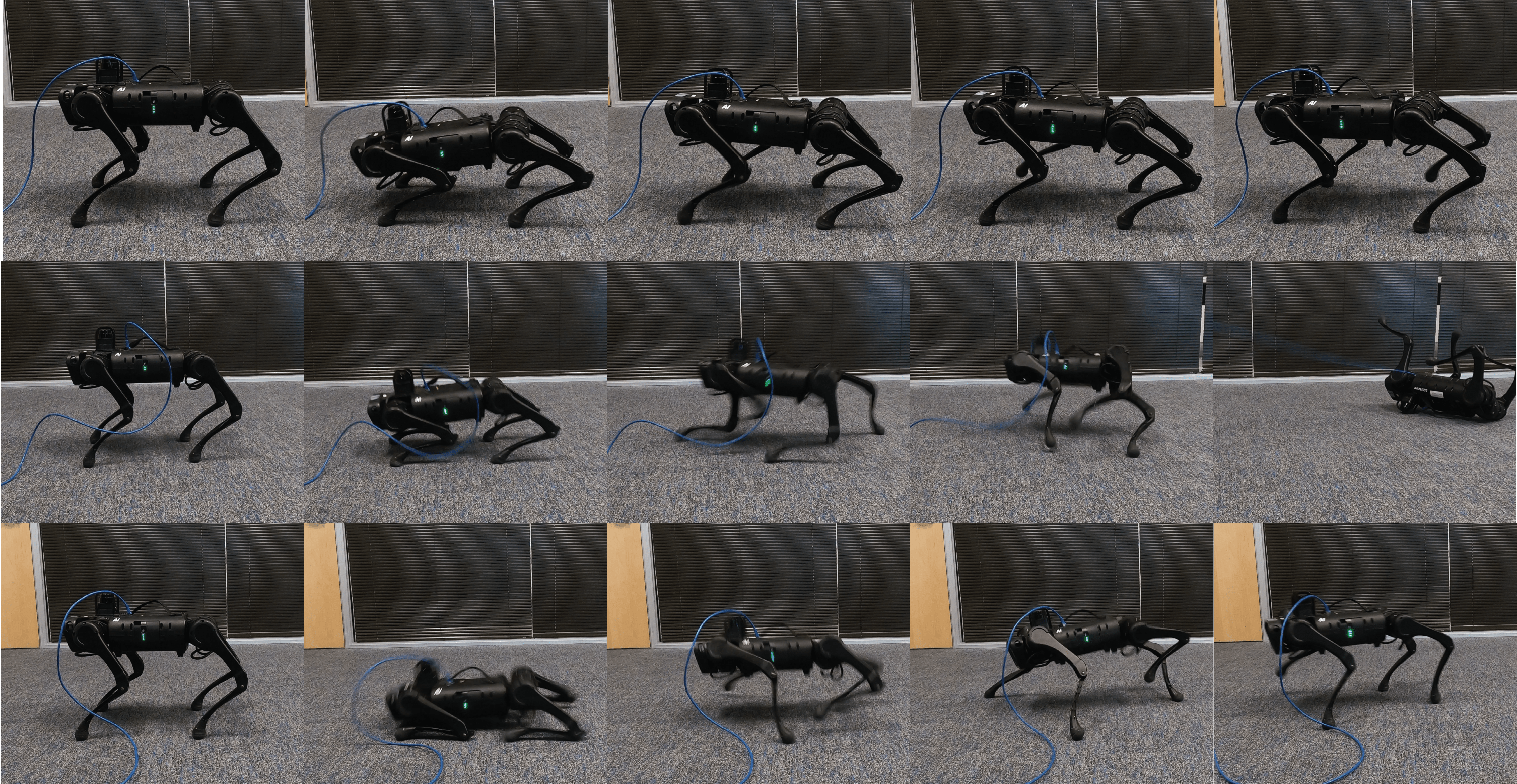}
    \caption{Turning off motor commands for 150 ms during the first step.
    HiLQR MPC (top row) creates a catching behavior and then goes back into the scheduled walk. 
    Instant QP sometimes tries to step to regulate velocity which destabilizes the robot (middle row). 
    Other times, Instant QP hits the ground  (bottom row), which stabilizes the body velocities and the robot is able to shoot its legs out in order to get back into the walk.}
    \label{fig:turn_off_torque}
\end{figure*}

The results of the motor-blocking physical robot experiment are shown in Fig. \ref{fig:turn_off_torque} and Table \ref{table::motor_block}.
Over five trials, HiLQR MPC was able to stabilize successfully after the motor block was released every time, while Instant QP was completely unstable $60$\% of the time and $40$\% of the time was able to stand up and walk after the body hit the ground.
Two unintended additional perturbations occurred in this experiment -- there is a consistent $10$ millisecond input delay on A1 and another perturbation caused by the state estimator.
The estimator relies on the kinematic information from the legs that are in contact to get a better estimate of the robot body.
However, when the motor commands were blocked, all the contact forces went close to zero, which resulted in a degraded estimate of the robot body until the legs made sufficient contact with the ground again.

HiLQR MPC was able to replan a stand trajectory in order to catch itself rather than sticking to the original plan of walking as Instant QP.
In the times that Instant QP successfully rejected the perturbation, the robot body actually hits the ground and the legs that are planned to be in stance apply enough standing force to get back up while the back right leg recirculates in order to counteract the backward velocity induced by getting back up. Since this relies on the body hitting the ground correctly and the back leg perfectly stabilizing the motion, it is a lot less reliable but is able to catch itself occasionally.

Similarly to the simulated experiments, HiLQR MPC outperforms the centroidal method (Instant QP) because HiLQR MPC does not have to adhere to a rigid gait schedule and can fluidly replan a new contact sequence to help stabilize the perturbation.
Although Instant QP utilizes the current contact information to inform which legs are in contact, the controller is trying its best to follow the scheduled gait sequence.
In this case, modifying the gait sequence from a walk to a stand is much more reliable.
HiLQR MPC is able to track the walking gait when appropriate but modify it to a stand if needed to catch the robot and seamlessly return to walking once the perturbation has been stabilized.
Having the ability to automatically modify the gait schedule to generate these stabilizing behaviors is important for a controller because the initial plan might not always be the best in the presence of disturbances.

\begin{table}[tb]
\caption{Motor blocking perturbation results over 5 trials.}
\centering
\vspace{.5em}
\begin{tabular}{l c c c}
\hline
Controller  & Success & Hit Ground & Uncontrolled\\
\hline
HiLQR MPC & $100$\%&$0$\% & $0$\%\\
Instant QP& $0$\%&$40$\% & $60$\%\\[2pt]
\hline
\label{table::motor_block}
\end{tabular}
\end{table}

\section{Discussion}
Allowing for varying contact sequences while planning for the full nonlinear dynamics of a robotic system is very difficult, but leads to more robust control.
In this work, we {present} Hybrid iLQR {and show how it can work as a} MPC controller, which can vary the contact sequence of the target trajectory as well as plan with the nonlinear dynamics. 
This is made possible by fixing gradient issues that occur when there are hybrid mode mismatches, using fast analytical derivatives of the contact dynamics, and parallelizing the line search in the forward pass.

In simulation, HiLQR MPC outperforms the state of the art centroidal motion planning technique (Convex MPC) for stabilizing perturbations 88\% success rate vs 50\% for large perturbations and 100\% vs 88\% for medium perturbations.
This is because HiLQR MPC is able to fully utilize the legs of the robot to help catch itself and can create more efficient and elegant solutions, needing fewer steps to recover.

HiLQR MPC is also able to run in real-time with some modifications to the hyperparameters and utilizing a hierarchical control structure where trajectories are sent to a lower level Hybrid LQR controller to track the hybrid trajectories planned by HiLQR MPC.
We are able to show for a motor blocking perturbation that HiLQR MPC is able to withstand this better than Instant QP, where HiLQR MPC succeeded for all trials and Instant QP could only stabilize 40\% of the time (and even then only after the robot body hit the ground).
The high success rate for HiLQR MPC is due to planning a reliable catching behavior, while Instant QP is continuously attempting to walk as best it can.

The code is currently implemented in Python while using several C++ libraries that utilize Python wrappers.
Improvements in the real-time application will be seen by further optimizing the code and implementing it in C++.
Overall, HiLQR MPC is a very modular MPC controller which can be run for any hybrid dynamical system of type Def. \ref{def:hs}.
Besides the hybrid dynamical systems definition, there are no restrictive simplifications that are made, which makes the controller generalizable to many different behaviors.
Future work will add additional constraints through Augmented Lagrangian \cite{howell2019altro} for obstacle avoidance and actuator constraints.



\bibliographystyle{IEEEtran}
\bibliography{references}

\begin{thebibliography}{10}
\providecommand{\url}[1]{#1}
\csname url@samestyle\endcsname
\providecommand{\newblock}{\relax}
\providecommand{\bibinfo}[2]{#2}
\providecommand{\BIBentrySTDinterwordspacing}{\spaceskip=0pt\relax}
\providecommand{\BIBentryALTinterwordstretchfactor}{4}
\providecommand{\BIBentryALTinterwordspacing}{\spaceskip=\fontdimen2\font plus
\BIBentryALTinterwordstretchfactor\fontdimen3\font minus
  \fontdimen4\font\relax}
\providecommand{\BIBforeignlanguage}[2]{{%
\expandafter\ifx\csname l@#1\endcsname\relax
\typeout{** WARNING: IEEEtran.bst: No hyphenation pattern has been}%
\typeout{** loaded for the language `#1'. Using the pattern for}%
\typeout{** the default language instead.}%
\else
\language=\csname l@#1\endcsname
\fi
#2}}
\providecommand{\BIBdecl}{\relax}
\BIBdecl

\bibitem{posa2014direct}
M.~Posa, C.~Cantu, and R.~Tedrake, ``A direct method for trajectory
  optimization of rigid bodies through contact,'' \emph{The International
  Journal of Robotics Research}, vol.~33, no.~1, pp. 69--81, 2014.

\bibitem{mordatch2012discovery}
I.~Mordatch, E.~Todorov, and Z.~Popovi{\'c}, ``Discovery of complex behaviors
  through contact-invariant optimization,'' \emph{ACM Transactions on Graphics
  (ToG)}, vol.~31, no.~4, pp. 1--8, 2012.

\bibitem{mombaur2009using}
K.~Mombaur, ``Using optimization to create self-stable human-like running,''
  \emph{Robotica}, vol.~27, no.~3, pp. 321--330, 2009.

\bibitem{diehl2006fast}
M.~Diehl, H.~G. Bock, H.~Diedam, and P.-B. Wieber, ``Fast direct multiple
  shooting algorithms for optimal robot control,'' in \emph{Fast motions in
  biomechanics and robotics}.\hskip 1em plus 0.5em minus 0.4em\relax Springer,
  2006, pp. 65--93.

\bibitem{von1999user}
O.~Von~Stryk, ``User’s guide for dircol: A direct collocation method for the
  numerical solution of optimal control problems,'' \emph{Lehrstuhl f{\"u}r
  H{\"o}here Mathematik und Numerische Mathematik, Technische Universit{\"a}t,
  M{\"u}nchen}, vol.~2, 1999.

\bibitem{posa2016optimization}
M.~Posa, S.~Kuindersma, and R.~Tedrake, ``Optimization and stabilization of
  trajectories for constrained dynamical systems,'' in \emph{IEEE International
  Conference on Robotics and Automation}, May 2016, pp. 1366--1373.

\bibitem{kelly2017introduction}
M.~Kelly, ``An introduction to trajectory optimization: How to do your own
  direct collocation,'' \emph{SIAM Review}, vol.~59, no.~4, pp. 849--904, 2017.

\bibitem{pardo2017hybrid}
D.~Pardo, M.~Neunert, A.~W. Winkler, R.~Grandia, and J.~Buchli, ``Hybrid direct
  collocation and control in the constraint-consistent subspace for dynamic
  legged robot locomotion.'' in \emph{Robotics: Science and Systems}, vol.~10,
  2017.

\bibitem{winkler2018gait}
A.~W. Winkler, C.~D. Bellicoso, M.~Hutter, and J.~Buchli, ``Gait and trajectory
  optimization for legged systems through phase-based end-effector
  parameterization,'' \emph{IEEE Robotics and Automation Letters}, vol.~3,
  no.~3, pp. 1560--1567, 2018.

\bibitem{manchester2021fast}
S.~Le~Cleac'h, T.~A. Howell, M.~Schwager, and Z.~Manchester, ``Fast
  contact-implicit model-predictive control,'' \emph{arXiv preprint
  arXiv:2107.0561}, 2021.

\bibitem{di2018dynamic}
J.~Di~Carlo, P.~M. Wensing, B.~Katz, G.~Bledt, and S.~Kim, ``Dynamic locomotion
  in the {MIT} {C}heetah 3 through convex model-predictive control,'' in
  \emph{IEEE/RSJ International Conference on Intelligent Robots and Systems},
  2018, pp. 1--9.

\bibitem{kim2019highly}
D.~Kim, J.~Di~Carlo, B.~Katz, G.~Bledt, and S.~Kim, ``Highly dynamic quadruped
  locomotion via whole-body impulse control and model predictive control,''
  \emph{arXiv preprint arXiv:1909.06586}, 2019.

\bibitem{da2020learning}
X.~Da, Z.~Xie, D.~Hoeller, B.~Boots, A.~Anandkumar, Y.~Zhu, B.~Babich, and
  A.~Garg, ``Learning a contact-adaptive controller for robust, efficient
  legged locomotion,'' in \emph{Conference on Robot Learning}, 16--18 Nov 2020,
  pp. 883--894.

\bibitem{xie2021glide}
Z.~Xie, X.~Da, B.~Babich, A.~Garg, and M.~v. de~Panne, ``Glide: Generalizable
  quadrupedal locomotion in diverse environments with a centroidal model,'' in
  \emph{Workshop on the Algorithmic Foundations of Robotics}.\hskip 1em plus
  0.5em minus 0.4em\relax Springer, 2022, pp. 523--539.

\bibitem{gehring2013control}
C.~Gehring, S.~Coros, M.~Hutter, M.~Bloesch, M.~A. Hoepflinger, and
  R.~Siegwart, ``Control of dynamic gaits for a quadrupedal robot,'' in
  \emph{IEEE international conference on Robotics and automation}, 2013, pp.
  3287--3292.

\bibitem{raibert1986legged}
M.~H. Raibert, \emph{Legged robots that balance}.\hskip 1em plus 0.5em minus
  0.4em\relax MIT press, 1986.

\bibitem{pratt2006capture}
J.~Pratt, J.~Carff, S.~Drakunov, and A.~Goswami, ``Capture point: A step toward
  humanoid push recovery,'' in \emph{IEEE-RAS International Conference on
  Humanoid Robots}, 2006, pp. 200--207.

\bibitem{mayne1966second}
D.~Mayne, ``A second-order gradient method for determining optimal trajectories
  of non-linear discrete-time systems,'' \emph{International Journal of
  Control}, vol.~3, no.~1, pp. 85--95, 1966.

\bibitem{li2004iterative}
W.~Li and E.~Todorov, ``Iterative linear quadratic regulator design for
  nonlinear biological movement systems.'' in \emph{International Conference on
  Informatics in Control, Automation and Robotics}, 2004, pp. 222--229.

\bibitem{tassa2012synthesis}
Y.~Tassa, T.~Erez, and E.~Todorov, ``Synthesis and stabilization of complex
  behaviors through online trajectory optimization,'' in \emph{IEEE/RSJ
  International Conference on Intelligent Robots and Systems}, 2012.

\bibitem{koenemann2015whole}
J.~Koenemann, A.~Del~Prete, Y.~Tassa, E.~Todorov, O.~Stasse, M.~Bennewitz, and
  N.~Mansard, ``Whole-body model-predictive control applied to the {HRP-2}
  humanoid,'' in \emph{IEEE/RSJ International Conference on Intelligent Robots
  and Systems}, 2015, pp. 3346--3351.

\bibitem{neunert2018whole}
M.~Neunert, M.~St{\"a}uble, M.~Giftthaler, C.~D. Bellicoso, J.~Carius,
  C.~Gehring, M.~Hutter, and J.~Buchli, ``Whole-body nonlinear model predictive
  control through contacts for quadrupeds,'' \emph{IEEE Robotics and Automation
  Letters}, vol.~3, no.~3, pp. 1458--1465, 2018.

\bibitem{dantec2021whole}
E.~Dantec, R.~Budhiraja, A.~Roig, T.~Lembono, G.~Saurel, O.~Stasse,
  P.~Fernbach, S.~Tonneau, S.~Vijayakumar, S.~Calinon, M.~Taix, and N.~Mansard,
  ``Whole body model predictive control with a memory of motion: Experiments on
  a torque-controlled {Talos},'' in \emph{IEEE International Conference on
  Robotics and Automation}, 2021, pp. 8202--8208.

\bibitem{li2020hybrid}
H.~Li and P.~M. Wensing, ``Hybrid systems differential dynamic programming for
  whole-body motion planning of legged robots,'' \emph{IEEE Robotics and
  Automation Letters}, vol.~5, no.~4, pp. 5448--5455, 2020.

\bibitem{mastalli2020crocoddyl}
C.~Mastalli, R.~Budhiraja, W.~Merkt, G.~Saurel, B.~Hammoud, M.~Naveau,
  J.~Carpentier, L.~Righetti, S.~Vijayakumar, and N.~Mansard, ``Crocoddyl: An
  efficient and versatile framework for multi-contact optimal control,'' in
  \emph{IEEE International Conference on Robotics and Automation}, 2020, pp.
  2536--2542.

\bibitem{mastalli2022agile}
C.~Mastalli, W.~Merkt, G.~Xin, J.~Shim, M.~Mistry, I.~Havoutis, and
  S.~Vijayakumar, ``Agile maneuvers in legged robots: a predictive control
  approach,'' \emph{arXiv preprint arXiv:2203.07554}, 2022.

\bibitem{li2021model}
H.~Li, R.~J. Frei, and P.~M. Wensing, ``Model hierarchy predictive control of
  robotic systems,'' \emph{IEEE Robotics and Automation Letters}, vol.~6,
  no.~2, pp. 3373--3380, 2021.

\bibitem{leine2004dynamics}
R.~Leine and H.~Nijmeijer, \emph{Dynamics and bifurcations of non-smooth
  mechanical systems}.\hskip 1em plus 0.5em minus 0.4em\relax Springer, 2004.

\bibitem{rijnen2015optimal}
M.~Rijnen, A.~Saccon, and H.~Nijmeijer, ``On optimal trajectory tracking for
  mechanical systems with unilateral constraints,'' in \emph{IEEE Conference on
  Decision and Control}, 2015, pp. 2561--2566.

\bibitem{aizerman1958determination}
M.~A. Aizerman and F.~R. Gantmacher, ``Determination of stability by linear
  approximation of a periodic solution of a system of differential equations
  with discontinuous right-hand sides,'' \emph{The Quarterly Journal of
  Mechanics and Applied Mathematics}, vol.~11, no.~4, 1958.

\bibitem{burden2018contraction}
S.~A. Burden, T.~Libby, and S.~D. Coogan, ``On contraction analysis for hybrid
  systems,'' 2018, arXiv:1811.03956.

\bibitem{paper:kong-ilqr-2021}
N.~J. Kong, G.~Council, and A.~M. Johnson, ``{iLQR} for piecewise-smooth hybrid
  dynamical systems,'' in \emph{IEEE Conference on Decision and Control},
  December 2021.

\bibitem{johnson2016hybrid}
A.~M. Johnson, S.~A. Burden, and D.~E. Koditschek, ``A hybrid systems model for
  simple manipulation and self-manipulation systems,'' \emph{The International
  Journal of Robotics Research}, vol.~35, no.~11, 2016.

\bibitem{Back_Guckenheimer_Myers_1993}
A.~Back, J.~M. Guckenheimer, and M.~Myers, ``A dynamical simulation facility
  for hybrid systems,'' in \emph{Hybrid Systems}, ser. Lecture Notes in
  Computer Science.\hskip 1em plus 0.5em minus 0.4em\relax Springer Berlin /
  Heidelberg, 1993, vol. 736.

\bibitem{lygeros2003dynamical}
J.~Lygeros, K.~H. Johansson, S.~N. Simic, J.~Zhang, and S.~S. Sastry,
  ``Dynamical properties of hybrid automata,'' \emph{IEEE Transactions on
  automatic control}, vol.~48, no.~1, pp. 2--17, 2003.

\bibitem{goebel2009hybrid}
R.~Goebel, R.~G. Sanfelice, and A.~R. Teel, ``Hybrid dynamical systems,''
  \emph{IEEE control systems magazine}, vol.~29, no.~2, pp. 28--93, 2009.

\bibitem{wehage1982dynamic}
R.~A. Wehage and E.~J. Haug, ``Dynamic analysis of mechanical systems with
  intermittent motion,'' \emph{Journal of Mechanical Design}, vol. 104, no.~4,
  pp. 778--784, 10 1982.

\bibitem{pfeiffer1996multibody}
F.~Pfeiffer and C.~Glocker, \emph{Multibody dynamics with unilateral
  contacts}.\hskip 1em plus 0.5em minus 0.4em\relax John Wiley \& Sons, 1996.

\bibitem{brogliato2002numerical}
B.~Brogliato, A.~Ten~Dam, L.~Paoli, F.~Ge{\'{}}not, and M.~Abadie, ``Numerical
  simulation of finite dimensional multibody nonsmooth mechanical systems,''
  \emph{Appl. Mech. Rev.}, vol.~55, no.~2, pp. 107--150, 2002.

\bibitem{stewart1996implicit}
D.~E. Stewart and J.~C. Trinkle, ``An implicit time-stepping scheme for rigid
  body dynamics with inelastic collisions and coulomb friction,''
  \emph{International Journal for Numerical Methods in Engineering}, vol.~39,
  no.~15, pp. 2673--2691, 1996.

\bibitem{anitescu1997formulating}
M.~Anitescu and F.~A. Potra, ``Formulating dynamic multi-rigid-body contact
  problems with friction as solvable linear complementarity problems,''
  \emph{Nonlinear Dynamics}, vol.~14, no.~3, pp. 231--247, 1997.

\bibitem{mayne1973differential}
D.~Q. Mayne, ``Differential dynamic programming--a unified approach to the
  optimization of dynamic systems,'' in \emph{Control and Dynamic
  Systems}.\hskip 1em plus 0.5em minus 0.4em\relax Elsevier, 1973, vol.~10, pp.
  179--254.

\bibitem{lantoine2012hybrid}
G.~Lantoine and R.~P. Russell, ``A hybrid differential dynamic programming
  algorithm for constrained optimal control problems. part 1: Theory,''
  \emph{Journal of Optimization Theory and Applications}, vol. 154, no.~2, pp.
  382--417, 2012.

\bibitem{isaacgym}
V.~Makoviychuk, L.~Wawrzyniak, Y.~Guo, M.~Lu, K.~Storey, M.~Macklin,
  D.~Hoeller, N.~Rudin, A.~Allshire, A.~Handa, and G.~State, ``Isaac {G}ym:
  High performance {GPU} based physics simulation for robot learning,'' in
  \emph{Conference on Neural Information Processing Systems, Datasets and
  Benchmarks Track}, 2021.

\bibitem{peachey2009parallel}
T.~C. Peachey, D.~Abramson, and A.~Lewis, ``Parallel line search,''
  \emph{Springer Optimization and Its Applications}, p. 369, 2009.

\bibitem{manchester2021planning}
B.~E. Jackson, K.~Tracy, and Z.~Manchester, ``Planning with attitude,''
  \emph{IEEE Robotics and Automation Letters}, vol.~6, no.~3, pp. 5658--5664,
  2021.

\bibitem{pinocchioweb}
J.~Carpentier, G.~Saurel, G.~Buondonno, J.~Mirabel, F.~Lamiraux, O.~Stasse, and
  N.~Mansard, ``The {P}inocchio {C}++ library -- {A} fast and flexible
  implementation of rigid body dynamics algorithms and their analytical
  derivatives,'' in \emph{International Symposium on System Integration}, 2019.

\bibitem{rijnen2019sensitivity}
M.~Rijnen, H.~L. Chen, N.~Van De~Wouw, A.~Saccon, and H.~Nijmeijer,
  ``Sensitivity analysis for trajectories of nonsmooth mechanical systems with
  simultaneous impacts: A hybrid systems perspective,'' in \emph{American
  Control Conference}.\hskip 1em plus 0.5em minus 0.4em\relax IEEE, 2019, pp.
  3623--3629.

\bibitem{burden2016event}
S.~A. Burden, S.~S. Sastry, D.~E. Koditschek, and S.~Revzen, ``Event--selected
  vector field discontinuities yield piecewise--differentiable flows,''
  \emph{SIAM Journal on Applied Dynamical Systems}, vol.~15, no.~2, pp.
  1227--1267, 2016.

\bibitem{council2022representing}
G.~Council, S.~Revzen, and S.~A. Burden, ``Representing and computing the
  b-derivative of the piecewise-differentiable flow of a class of nonsmooth
  vector fields,'' \emph{Journal of Computational and Nonlinear Dynamics},
  vol.~17, no.~9, p. 091004, 2022.

\bibitem{shampine2003solving}
L.~F. Shampine, I.~Gladwell, L.~Shampine, and S.~Thompson, \emph{Solving {ODEs}
  with {MATLAB}}.\hskip 1em plus 0.5em minus 0.4em\relax Cambridge University
  Press, 2003.

\bibitem{lussier2011landing}
A.~Lussier~Desbiens, A.~T. Asbeck, and M.~R. Cutkosky, ``Landing, perching and
  taking off from vertical surfaces,'' \emph{The International Journal of
  Robotics Research}, vol.~30, no.~3, pp. 355--370, 2011.

\bibitem{bledt2018contact}
G.~Bledt, P.~M. Wensing, S.~Ingersoll, and S.~Kim, ``Contact model fusion for
  event-based locomotion in unstructured terrains,'' in \emph{IEEE
  International Conference on Robotics and Automation}, 2018, pp. 4399--4406.

\bibitem{howell2019altro}
T.~A. Howell, B.~E. Jackson, and Z.~Manchester, ``Altro: A fast solver for
  constrained trajectory optimization,'' in \emph{IEEE/RSJ International
  Conference on Intelligent Robots and Systems}, 2019, pp. 7674--7679.

\end{thebibliography}
\newpage
\section{Biography}
\vspace{-11pt}
\begin{IEEEbiography}[{\includegraphics[width=1in,height=1.25in,clip,keepaspectratio]{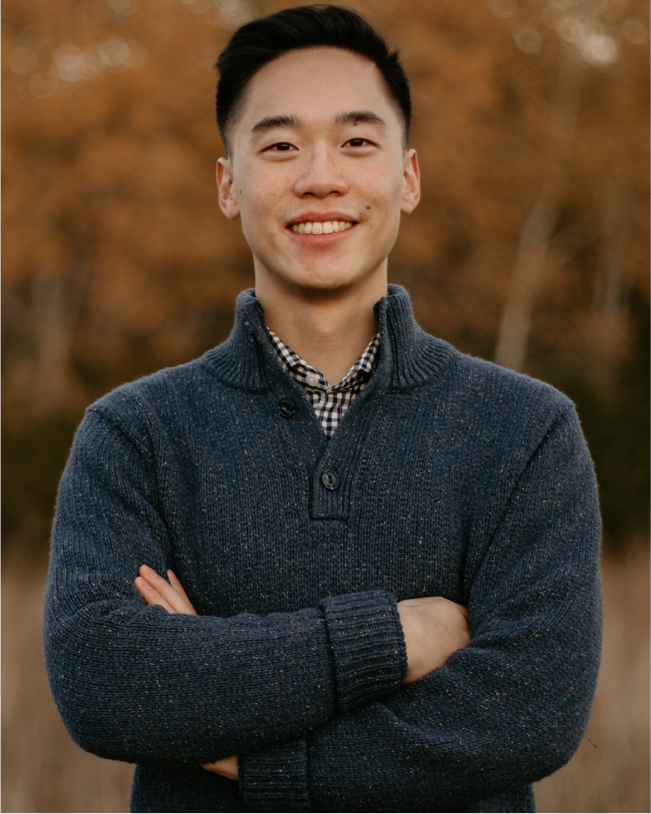}}]{Nathan Kong}
received his B.S. degree in mechanical engineering from the University of Minnesota Twin Cities,
Minneapolis, MN, USA, in 2018 and his Ph.D. degree in mechanical engineering at Carnegie
Mellon University, Pittsburgh, PA, USA, in 2022. He is currently an Applied Scientist at Amazon. His research
interests are in hybrid dynamical systems, legged
locomotion, trajectory optimization, control, and path planning. 
\end{IEEEbiography}
\vspace{11pt}
\begin{IEEEbiography}[{\includegraphics[width=1in,height=1.25in,clip,keepaspectratio]{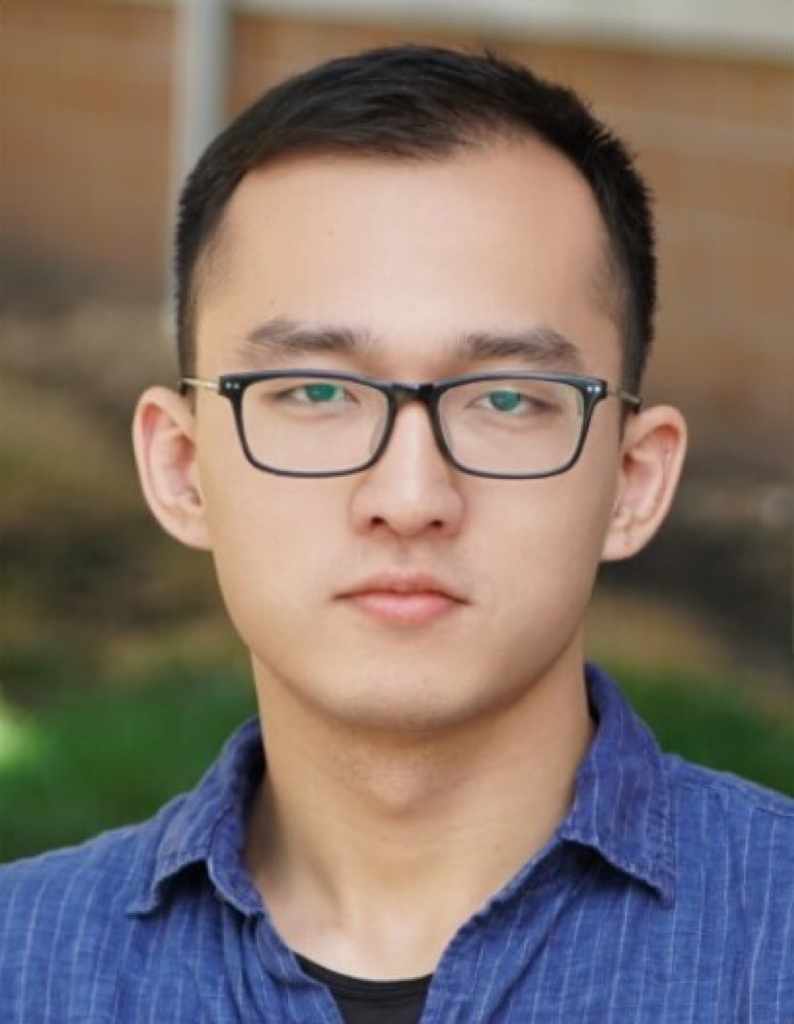}}]{Chuanzheng Li}
(S’17-S’19) received his B.S. degree in mechatronics from Zhejiang University, Hangzhou, China, in 2014 and completed his Ph.D. in mechanical engineering at the University of Illinois at Urbana-Champaign, Champaign, IL, USA, in 2022. His research interests lie at the intersection of mechatronic design, model-based and RL-bassed control of legged robots. He is currently working as Staff Control Engineer at PX Robotics Inc., focusing on the real-time control of legged locomotion.
\end{IEEEbiography}
\vspace{-11pt}

\begin{IEEEbiography}[{\includegraphics[width=1in,height=1.25in,clip,keepaspectratio]{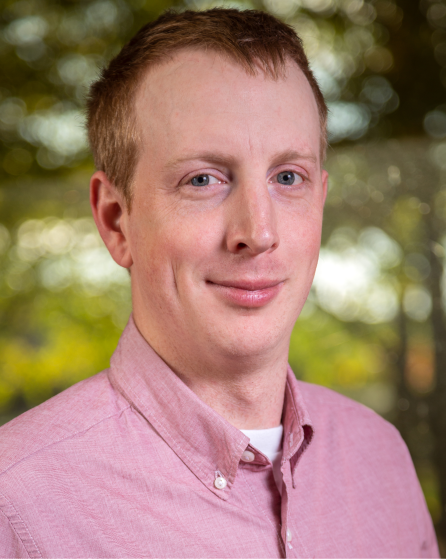}}]{George Council}
(S'10 M'20) was born in Laurel, Maryland in 1988. He received his B.S. of electrical engineering from Montana State University, Bozeman, MT, USA, in 2012, and subsequently completed his Ph.D degree in electrical engineering with the BIRDS Lab at the University of Michigan, Ann Arbor, MI USA, in 2020. He is currently an ADAS engineer for Ford Motor Company, while previously he was a postdoctoral fellow at Carnegie Mellon University. His research interests include legged robotics, hybrid dynamics and control, data-driven methods in applied mathematics, and their integration as a unified perspective. Dr. Council is a member of the IEEE, as well as Society for Industrial and Applied Mathematics (SIAM).
\end{IEEEbiography}
\vspace{-11pt}
\begin{IEEEbiography}[{\includegraphics[width=1in,height=1.25in,clip,keepaspectratio]{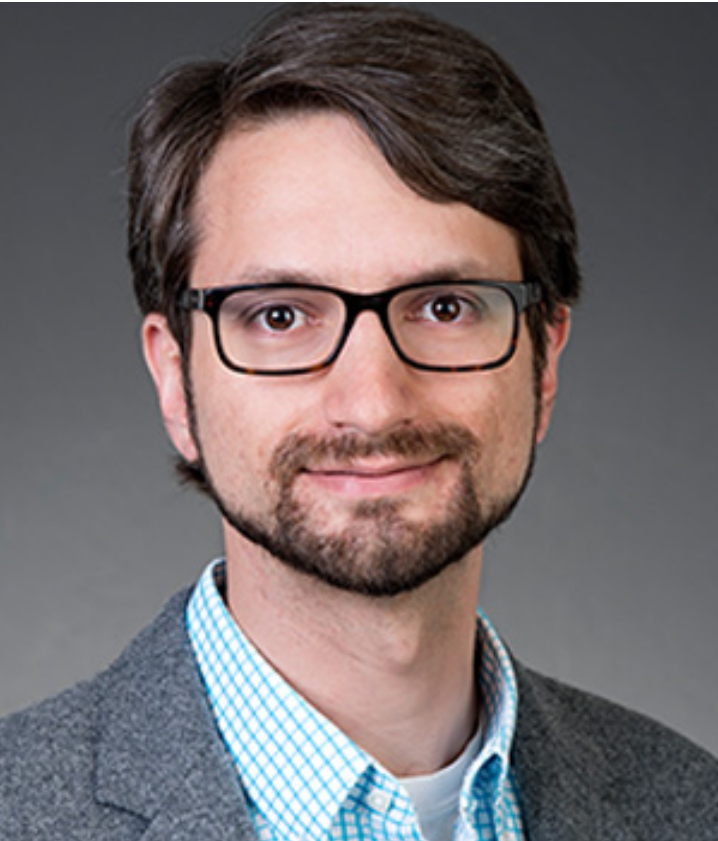}}]{Aaron Johnson} (S'06–M'14-SM'19)
received the B.S. degree in electrical and computer engineering from Carnegie Mellon
University, Pittsburgh, PA, USA, in 2008 and the Ph.D.
degree in electrical and systems engineering from the
University of Pennsylvania, Philadelphia, PA, USA, in
2014. 

He is currently an Associate Professor of Mechanical Engineering at Carnegie Mellon University,
with appointments in the Robotics Institute and Electrical and Computer Engineering Department. He was
previously a Postdoctoral Fellow at Carnegie Mellon
University and the University of Pennsylvania. 
His research interests include legged locomotion, hybrid dynamical systems, robust control, and bioinspired robotics. 

Prof. Johnson received the NSF CAREER award
in 2020 and the ARO Young Investigator Award in 2018. He and his students won the Best Workshop Paper at the 2022 ICRA Workshop on Legged Robotics. He has been an Associate Editor for the IEEE International Conference on Robotics and Automation since 2017.

\end{IEEEbiography}

\end{document}